\documentclass[11pt,letterpaper]{article}

\usepackage[margin=1in]{geometry}
\usepackage[T1]{fontenc}
\usepackage{lmodern}
\usepackage{microtype}
\usepackage[hyphens]{url}
\usepackage{graphicx}
\usepackage{booktabs}
\usepackage{amsmath,amssymb}
\usepackage{tabularx}
\usepackage{multirow}
\usepackage{natbib}
\usepackage{caption}
\usepackage{xcolor}
\usepackage[hidelinks]{hyperref}

\newcommand{\vraw}{\mathbf{v}_{\mathrm{raw}}}
\newcommand{\h}{\mathbf{h}}

\title{What Does Activation Steering Control?\\
Attribution Across Answer Encodings and Output-Sensitive Subspaces}
\author{
Zhiwei Gao 
\qquad
Shaowen Peng 
\qquad
Shoko Wakamiya
\qquad
Eiji Aramaki \\[0.5em]
Nara Institute of Science and Technology
}
\date{}

\begin{document}
\maketitle

\begin{abstract}
Activation steering is often evaluated under the answer encoding used to construct the direction. A reported gain may reflect the intended judgment or compatibility with answer identifiers seen during construction. We introduce \emph{Cross-Encoding Steering Evaluation}, which freezes an intervention while re-encoding answers to the same held-out items. On NormBank, after A/B/C identifiers are reassigned, contrastive activation addition (CAA) induces larger target-versus-source score changes for the extraction indices than for the semantic labels under the new mapping. We call this extraction-index following. Varying identifier vocabulary (A/B/C, X/Y/Z, or 1/2/3) and row order shows that the effect tracks extraction index rather than row position. After matching direction norms across layers, extraction-index following emerges mainly at later depths. A low-rank output-sensitive component containing 15.4\% of the direction's squared norm retains 96.3\% of this effect. An Inference-Time Intervention (ITI)-style method also favors extraction-index over semantic-label following on NormBank in three models. In aggregate, MNLI favors extraction-index following, whereas Social Chemistry 101 (SC101) favors semantic-label following. Multiple-choice and open-ended evaluations can yield different behavioral conclusions. Thus, a steering gain under one answer encoding does not by itself identify what the intervention controls.
\end{abstract}

\section{Introduction}

Activation steering changes large language model (LLM) behavior without updating weights. Contrastive activation addition (CAA) averages hidden-state differences between positive and negative examples and injects the resulting direction at inference time \cite{panickssery2024caa}. Related methods intervene on residual streams or attention heads \cite{zou2023representation,li2023iti}. Researchers have applied these methods to moral profiles, values, and cultural adaptation \cite{tlaie2024moralcompass,sauter2026rules,jin2025conva,dang2026cultural,dang2026scenario,yang2026neva}. This work leaves a basic evidential question: when an intervention improves its reported score, what does it actually control?

Consider a direction extracted with A = taboo, B = normal, and C = expected. A gain toward C could reflect the semantic class \emph{expected}, the option identifier ``C'', or its displayed row. To distinguish these possibilities, we keep the direction fixed and change only the answer encoding. For example, if evaluation instead uses A = expected, B = taboo, and C = normal, then semantic-label following favors the option currently assigned to expected (A), whereas extraction-index following favors C. Cross-encoding evaluation exploits this disagreement without changing the intervention itself.

To describe these alternatives, we use three terms throughout the paper. \emph{Semantic-label following} tracks the target semantic label under the current test encoding: it favors whichever identifier is currently assigned to that label. \emph{Extraction-index following} tracks the target's identifier index under the extraction encoding; across identifier vocabularies we align A/X/1, B/Y/2, and C/Z/3. \emph{Extraction-row following} instead tracks whichever identifier is currently displayed in the target's extraction-time row.

\begin{figure*}[!t]
    \centering
    \includegraphics[width=\textwidth]{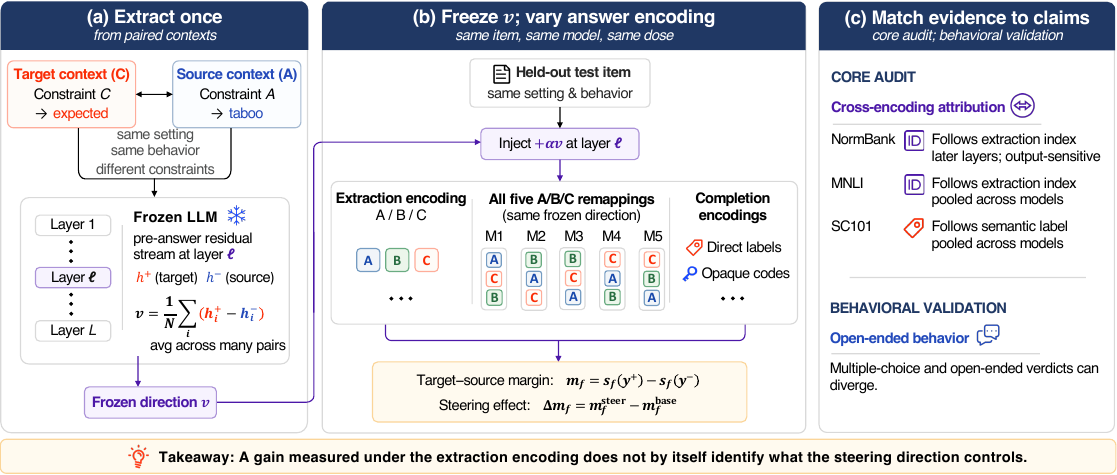}
\caption{
Cross-Encoding Steering Evaluation. We extract a residual-stream CAA direction from paired target/source contexts, with the target denoting the endpoint the direction is constructed to favor and the source the endpoint subtracted from it. We then counterfactually re-render the answer scaffold, holding the underlying test item, model, direction, layer, position, and dose fixed.
In Panel~(b), each remapping column shows, from top to bottom, the option identifiers assigned to \textit{taboo}, \textit{normal}, and \textit{expected}; the displayed row order is fixed in these five examples.
Here, $s_f(y)$ is the mean per-token log likelihood of the sequence expressing semantic label $y$ under answer encoding $f$, and $\Delta m_f$ is the steered-minus-base change in the target--source margin.
Because letter and completion encodings use different scoring scales, margins are compared only within the same encoding.
}
\label{fig:method-overview}
\end{figure*}

We use \emph{answer encoding} to mean how semantic labels are mapped to and expressed as candidate answers, including label-to-identifier mapping, identifier vocabulary, row order, and completion format.  
\emph{Cross-Encoding Steering Evaluation} re-renders the answer scaffold for the same held-out items while holding the intervention fixed (Figure~\ref{fig:method-overview}). Remapping and factorial audits distinguish the three forms of following; layer, position, and subspace interventions localize the effect. Human-validated open-ended responses test generative control separately.

\paragraph{Main findings.} On NormBank, the extraction-index effect exceeds the semantic-label effect under all five remappings, so extraction-encoding gain alone is not diagnostic of semantic-label control. After norm matching, the effect emerges at later tested layers; a low-rank output-sensitive component containing 15.4\% of the direction's squared norm retains 96.3\% of the effect, revealing strong concentration. Inference-Time Intervention (ITI) reproduces this ordering, and MNLI does so in aggregate; Qwen reverses on MNLI, while Social Chemistry 101 (SC101) favors semantic-label following. Open-ended tests show that MCQ scores and generative behavior can yield different conclusions.

We organize the study around one core attribution question and one behavioral validation:
\begin{itemize}
    \item \textbf{Core audit: Attribution and localization.} When answer encoding changes, does a fixed effect track the semantic label, extraction index, or row, and where does it emerge?
    \item \textbf{Behavioral validation: Open-ended behavior.} Do multiple-choice (MCQ) scores and open generations support the same conclusion?
\end{itemize}

\paragraph{Contributions.} We introduce a fixed-intervention attribution framework that separates semantic-label, extraction-index, and extraction-row following under counterfactual answer encodings. We show that extraction-index following emerges at late depths and concentrates in output-sensitive components, while attribution varies across models, tasks, and evaluation granularities.

\section{Related Work}
\paragraph{Steering evaluation.} CAA and representation engineering inject contrastive activation directions \cite{panickssery2024caa,zou2023representation}; ITI selects and edits probe-informative attention heads \cite{li2023iti}. Prior work documents input sensitivity, spurious dependence, and out-of-distribution brittleness \cite{tan2024steeringreliability}. Complementary reliable-evaluation work emphasizes context-matched tasks, likelihood-aware metrics, standardized comparisons, and explicit baselines \cite{pres2024reliable}. Evaluations may also disagree across behavioral granularities \cite{xu2026steereval}. Multiple-choice studies vary identifiers or verbalizers to test unmodified models \cite{zheng2023mcqselectors,li2024verbalizer}. 
Recent work shows that steering efficacy depends strongly on activation-source selection, with execution-boundary states yielding particularly effective signals \cite{ye2026steering},
and that output-stage scaffold features can dominate decision logits despite preserved task-relevant representations in unmodified models \cite{fraile2026internal}. We complement these findings by freezing the intervention and counterfactually varying answer encoding to attribute the intervention-induced change itself.

\paragraph{Output-sensitive subspaces.} Intermediate prediction, representation geometry, and gradient attribution connect hidden states to model outputs \cite{belrose2023tunedlens,park2023linear,kramar2024atp}. DecodeShare identifies a low-dimensional task-shared decode-time subspace and causally decomposes steering directions relative to it \cite{shao2026decodeshare}. Our analysis adapts this projection--intervention logic to a different attribution target: we derive an identifier-readout-specific local subspace from option-logit gradients and test whether the projection of a CAA direction onto that subspace carries extraction-index following. Behaviorally equivalent steering vectors pose a separate non-identifiability problem \cite{venkatesh2026nonidentifiability}; our audit instead holds each direction fixed and attributes its induced change across counterfactual answer encodings.

\section{Method}

\subsection{Contrastive Activation Steering}

We write $x$ for a rendered prompt prefix ending before the scored candidate. For each contrastive pair $i$, $x_i^+$ denotes the \emph{target} endpoint, toward which the CAA direction is constructed to point, and $x_i^-$ the \emph{source} endpoint subtracted from it.
The residual-stream state at block $\ell$ and position $p$ is $h_{\ell,p}(x)\in\mathbb{R}^d$, where $d$ is the hidden size, and $\Delta a = a^{\mathrm{steer}}-a^{\mathrm{base}}$.
For contrast $c$ with $N_c$ pairs, the canonical direction extracted at $p_{\mathrm{ext}}$, denoted $\vraw$, is
\begin{equation}
\begin{aligned}
\vraw^{(c,p_{\mathrm{ext}})}
&=\frac{1}{N_c}\sum_{i=1}^{N_c}
\Bigl(\h_{\ell,p_{\mathrm{ext}}}(x_i^+) - \h_{\ell,p_{\mathrm{ext}}}(x_i^-)\Bigr).
\end{aligned}
\end{equation}
At inference, a forward hook adds the direction at injection position $p_{\mathrm{inj}}$ in the same block:
\begin{equation}
\begin{aligned}
\h_{\ell,p_{\mathrm{inj}}}(x)
&\leftarrow \h_{\ell,p_{\mathrm{inj}}}(x) +\alpha\vraw^{(c,p_{\mathrm{ext}})}.
\end{aligned}
\end{equation}
The primary audit extracts and injects at the pre-answer position. We fix the direction, block, position, and strength $\alpha$ before evaluating any test answer encoding. $L_2$-matched Gaussian vectors serve as perturbation controls.

\subsection{Cross-Encoding Evaluation}
\label{sec:fixed-remapping}
The audit holds one direction fixed and changes only the answer encoding. An encoding specifies how semantic labels are mapped to option identifiers and how candidate answers are represented for scoring. We test all six A/B/C assignments, direct label words, and opaque codewords. Each encoding has its own output sequences and margin scale, so raw-margin comparisons remain within an encoding.

Let $y^\pm$ be the target/source labels and $T_f(y)=(t_1,\ldots,t_K)$ their $K$-token rendering under encoding $f$. We use mean sequence log likelihood
\begin{equation}
\begin{aligned}
s_f(y\mid x)
&=\frac{1}{K}\sum_{j=1}^{K}
\log P(t_j\mid x,t_{<j}),\\
m_f(x)&=s_f(y^+\mid x)-s_f(y^-\mid x).
\end{aligned}
\end{equation}
Here $t_{<j}=(t_1,\ldots,t_{j-1})$. Positive $\Delta m_f$ indicates movement toward the target under the current test mapping.

For A/B/C mapping $\pi$, let $s_\pi(o\mid x)$ be the mean-token log likelihood of identifier $o$. Let $\pi_0$ denote the extraction mapping, and let $k_0^\pm$ be the identifier indices assigned to the target/source semantic labels under $\pi_0$. Let $o_k$ denote the A/B/C identifier at index $k$, so that $o_1=A$, $o_2=B$, and $o_3=C$. For vocabulary $z$, let $o_{z,k}$ denote the identifier occupying index $k$.
For example, if the extraction-time target identifier is ``C'', then its extraction index is $k_0^+=3$. Under the aligned vocabularies A/B/C, X/Y/Z, and 1/2/3, this same extraction index is instantiated by C, Z, and the token ``3'', respectively. The extraction index therefore tracks the identifier index defined at extraction time, not the currently assigned semantic label or displayed row.

Let
$\Delta s_\pi=s_\pi^{\mathrm{steer}}-s_\pi^{\mathrm{base}}$.
We define the extraction-index effect for the A/B/C remapping
audit as
\begin{equation}
\Delta m_{\mathrm{idx},\pi}(x)
=
\Delta s_\pi(o_{k_0^+}\mid x)
-
\Delta s_\pi(o_{k_0^-}\mid x).
\end{equation}
The semantic-label effect $\Delta m_{\mathrm{sem},\pi}(x)=\Delta m_{\pi}(x)$ instead follows the target and source semantic labels under the current test mapping. 
Their difference defines extraction-index advantage:
\begin{equation}
A_{\mathrm{idx},\pi}(x)
=
\Delta m_{\mathrm{idx},\pi}(x)
-
\Delta m_{\mathrm{sem},\pi}(x).
\end{equation}
Thus, $A_{\mathrm{idx},\pi}>0$ means that the paired extraction-index score change exceeds the semantic-label score change. We report both the extraction-index effect and extraction-index advantage.
The effects coincide under the extraction mapping and separate after remapping, using the same steered and base passes. Because those indices may be assigned to different semantic classes after remapping, a positive extraction-index effect need not indicate movement toward the current target label.

Random-adjusted estimates subtract $L_2$-matched random-direction effects, controlling for general encoding sensitivity to residual perturbations, and weight model--contrast strata equally. Extraction-index advantage is formed per pair before adjustment. Standardized sensitivity analyses appear in Appendix~\ref{sec:statistical-details}; unstandardized effects are interpreted within their encoding.

\subsection{Factorial Attribution of Semantic Label, Extraction Index, and Extraction Row}
Letter remapping alone leaves A/B/C attached to fixed rows. 
To separate identifier index from displayed row, the factorial audit crosses six semantic mappings, three aligned identifier vocabularies (A/B/C, X/Y/Z, and 1/2/3; A/X/1, B/Y/2, C/Z/3), and all six row orders. Vocabulary and row order vary independently while the extraction-index alignment defined above is preserved, yielding 108 conditions per frozen direction and test pair.

The factorial audit scores each candidate identifier by its mean-token log likelihood and computes steered-minus-base changes in the semantic-label, extraction-index, and extraction-row margins. For example, after replacing A/B/C with X/Y/Z and reordering the rows, the semantic-label margin follows the identifier currently assigned to the target and source semantic labels, the extraction-index margin follows the identifier at the same extraction index (e.g., C$\rightarrow$Z), and the extraction-row margin follows the original displayed position. 
Subtracting the extraction-row effect separates identifier index from row position. Unlike the random-adjusted letter audit, factorial effects are unadjusted steered-minus-base changes. We check unsteered accuracy under every mapping, and each model--mapping--identifier-vocabulary condition must pass an 80\% mapping-comprehension check. Estimates average conditions within each pair and weight contrasts equally. The group-cluster bootstrap jointly resamples complete setting--behavior groups across contrasts.

\subsection{Position and Output-Sensitive Subspace Localization}
\label{sec:readout-geometry}
We localize the identifier-linked effect across model depth, extraction position, and output-sensitive subspaces to test when it emerges, whether it depends on the answer scaffold, and which output-sensitive component carries it. A depth sweep evaluates the extraction-index effect and extraction-index advantage at 50\%, 62.5\%, 75\%, and 87.5\% of each model after rescaling every model--contrast direction to its 75\%-depth $L_2$ norm. The pre-answer state is taken after the model has seen the question, option identifiers, label descriptions, and output instruction, whereas the scenario-end state precedes this answer scaffold. We first compare otherwise identical directions extracted at the scenario end and at the pre-answer position while injecting both at the same pre-answer site. Norm matching tests whether the loss is explained by direction magnitude, while matched-position injection tests whether it arises from moving a scenario-end direction to the pre-answer site.

To identify which part of a pre-answer direction carries extraction-index following, we estimate a local identifier-readout subspace. 
For prompt $x$ and identifier pair $(o_a,o_b)$, let $p_{\mathrm{pre}}$ denote the pre-answer position and $z(o)$ the next-token logit for $o$. We define
\begin{equation}
\mathbf g_{x,a,b}
=\nabla_{\h_{\ell,p_{\mathrm{pre}}}(x)}
\left[z(o_a)-z(o_b)\right].
\end{equation}
For a fixed prompt, $\mathbf g_{x,a,b}$ points in the hidden-state direction of steepest first-order increase in the logit preference for $o_a$ over $o_b$ under an $L_2$-normalized perturbation.
Within each model, let $\mathbf G\in\mathbb R^{n_g\times d}$ stack the row-normalized training gradients from all three contrasts, and write its SVD as $\mathbf G=\mathbf L\mathbf\Sigma\mathbf R^\top$. The leading right singular vectors summarize the dominant identifier-logit sensitivities
across training prompts. We define
$\mathbf U_r=\mathbf R_{[:,1:r]}\in\mathbb R^{d\times r}$,
where $r\in\{2,4,8,16\}$ is the smallest rank for which the mean fraction of squared validation-gradient norm captured by $\mathbf U_r$ is at least 90\%.
Thus, the columns of $\mathbf U_r$ are orthonormal hidden-state directions that locally change option-identifier logit differences. Projection--residual interventions then test whether this subspace carries the extraction-index effect; test prompts are reserved for evaluation. We decompose
\begin{equation}
\begin{gathered}
\vraw = \mathbf v_{\parallel}+\mathbf v_{\perp},\\
\mathbf v_{\parallel} = \mathbf U_r\mathbf U_r^\top\vraw,
\qquad
\mathbf v_{\perp} = (\mathbf I-\mathbf U_r\mathbf U_r^\top)\vraw.
\end{gathered}
\end{equation}
The projection--residual decomposition asks whether the measured extraction-index effect is concentrated in a small component of the CAA direction or merely follows where most of the direction's norm lies. We therefore compare each component's extraction-index effect with its share of the full direction's squared $L_2$ norm. For a component $\mathbf u$, we define its energy fraction as \(E(\mathbf u)=\|\mathbf u\|_2^2/\|\vraw\|_2^2\). A component with low energy fraction but high effect retention indicates concentration of the measured effect in that subspace.

We define effect retention as the component's extraction-index effect divided by that of the full canonical CAA direction, computed on the same evaluation scale.
We intervene with the projection, residual, and their norm-matched variants, alongside ten covariance-matched random controls. Validation prompts assess whether the local Jacobian's first-order predictions agree with the observed intervention effects.

A direct output-sensitive baseline averages normalized local logit-margin gradients from strict-split training prompts, matches the result to the canonical CAA norm, and freezes it before test evaluation. This tests whether a simple output-sensitive controller reproduces the extraction-index-over-semantic-label ordering. For cross-vocabulary transfer, we freeze the A/B/C-derived basis, projection, residual, layer, and dose, then evaluate the same test pairs with A/B/C, X/Y/Z, and numeric identifiers. The primary comparison uses model--mapping cells that pass the mapping-comprehension check under all three vocabularies. Further results appear in Appendix~\ref{sec:cross-vocab-transfer}.

\subsection{Scope and Behavioral Validation}
We next test whether the attribution extends beyond canonical NormBank CAA. For cross-method replication, an ITI-style intervention \cite{li2023iti} ranks attention heads by validation probe accuracy and adds normalized target-minus-source class-mean directions at the selected heads. Validation selects the layers, head count, and strength before mapping evaluation. A prespecified positive-gain criterion retains only models whose validation-selected mean gain under the extraction encoding is positive; the wrong-sign control keeps the selected heads and magnitude fixed but reverses every direction. For cross-task scope, MNLI and SC101 repeat the frozen six-mapping audit on same-premise and action-level pairs, respectively.

The open-ended behavior test reproduces public CAA hallucination, refusal, and sycophancy directions on Llama-2-7B/13B-Chat \cite{panickssery2024caa} and applies them to the published open prompts. Three fixed-rubric LLM judges score 918 generations, using the per-item median for paired effects. Two blinded annotators and a third adjudicator validate a prespecified 180-response subset spanning every model--behavior and dose condition.

Across these analyses, $L_2$-matched random directions control for answer-encoding-specific perturbations, and opaque-key checks verify temporary-key decoding. Primary results average five fixed Gaussian controls and weight model--contrast strata equally. ITI and cross-task breakdowns appear in Appendix~\ref{sec:supp-scope}.

\begin{table*}[!t]
\centering
\footnotesize
\setlength{\tabcolsep}{2pt}
\begin{tabular}{@{}llrrrr@{}}
\toprule
\multicolumn{1}{c}{Labels assigned to A/B/C}
&
\multicolumn{1}{c}{Type}
&
\multicolumn{1}{c}{\begin{tabular}[c]{@{}c@{}}Extraction-index\\effect\end{tabular}}
&
\multicolumn{1}{c}{\begin{tabular}[c]{@{}c@{}}Semantic-label\\effect\end{tabular}}
&
\multicolumn{1}{c}{\begin{tabular}[c]{@{}c@{}}Extraction-index\\advantage\end{tabular}}
&
\multicolumn{1}{c}{Positive pairs}
\\
\midrule
T/N/E                    & Extraction & 4.63 {[}4.59, 4.67{]}                                          & 4.63 {[}4.59, 4.67{]}                                          & 0.00                                                              & 100.0\%        \\ \midrule
E/T/N                    & 3-cycle    & 3.00 {[}2.98, 3.03{]}                                          & $-0.33$ {[}$-0.38$, $-0.28${]}                                 & 3.33 {[}3.28, 3.38{]}                                             & 93.8\%         \\
N/E/T                    & 3-cycle    & 3.07 {[}3.04, 3.10{]}                                          & $-0.97$ {[}$-1.02$, $-0.92${]}                                 & 4.04 {[}3.99, 4.10{]}                                             & 97.6\%         \\
N/T/E                    & Swap       & 3.56 {[}3.53, 3.59{]}                                          & 0.56 {[}0.53, 0.59{]}                                          & 3.00 {[}2.96, 3.05{]}                                             & 96.9\%         \\
T/E/N                    & Swap       & 4.13 {[}4.09, 4.17{]}                                          & 2.18 {[}2.14, 2.22{]}                                          & 1.95 {[}1.91, 1.98{]}                                             & 97.0\%         \\
E/N/T                    & Swap       & 3.19 {[}3.16, 3.22{]}                                          & 0.02 {[}$-0.01$, 0.05{]}                                       & 3.17 {[}3.12, 3.22{]}                                             & 94.4\%         \\ 
\bottomrule
\end{tabular}
\caption{Residual-stream CAA on the strict split. E/T/N denote expected/taboo/normal. Extraction-index and semantic-label effects are random-adjusted changes on the common A/B/C mean-token log-likelihood-margin scale; extraction-index advantage is their pairwise difference (extraction-index minus semantic-label effect). Intervals resample complete setting--behavior groups; ``positive pairs'' is the pair-level sign rate for the extraction-index effect.}
\label{tab:letter-permutations}
\end{table*}

\section{Experimental Setup}

\paragraph{Datasets.} We use NormBank as the primary controlled audit because its \textit{taboo} (T), \textit{normal} (N), and \textit{expected} (E) labels vary with contextual constraints for matched settings and behaviors, enabling the ordered contrasts T$\rightarrow$N, T$\rightarrow$E, and N$\rightarrow$E while reducing content confounding \cite{ziems2023normbank}.
MNLI supplies a non-norm three-label task with same-premise pairs \cite{williams2018broad}; SC101 supplies action-level \textit{bad}/\textit{ok}/\textit{good} pairs that are not context matched \cite{forbes2020socialchemistry}. The public CAA behaviors extend the evaluation to generation.

\paragraph{Pairs and splits.} NormBank endpoints are matched within contrast and complete \texttt{setting + behavior} groups are assigned to train, validation, or test, with no shared pairs, groups, or endpoints. CAA evaluates up to 512 test pairs per contrast; the factorial audit applies all 108 conditions. MNLI analogously splits by normalized premise. 

\paragraph{Models.} NormBank, MNLI, and SC101 use Qwen2.5-7B-Instruct, Llama-3.1-8B-Instruct, Mistral-7B-Instruct-v0.3, and Gemma-2-9B-IT \cite{qwen2024qwen25,dubey2024llama3,jiang2023mistral,gemma2024gemma2}. CAA reports all four; ITI reports the three models retained by its prespecified positive-gain criterion. The public protocol uses Llama-2-7B/13B-Chat.

\paragraph{Locked interventions.} Primary NormBank and MNLI CAA use zero-based blocks 20, 23, 23, and 31 (Qwen/Llama/Mistral/Gemma), each nearest 75\% depth, with fixed $\alpha=.8$. The layer audit also evaluates 50\%, 62.5\%, and 87.5\% depth with directions norm-matched within each model and contrast to the 75\%-depth reference. ITI selects its layer, heads, and strength on validation data; SC101 uses previously selected middle layers and $\alpha=1$. Test answer encodings never change the extracted direction or any hyperparameter.

\paragraph{Scoring.} Cross-encoding and factorial comparisons use mean-token log-likelihood margins within each encoding. Layer, position, and subspace localization use normalized option-choice probability margins (target probability minus source probability) on a separate scale. Effect retention divides a component's equal-stratum mean extraction-index effect by canonical CAA. Intervals resample complete setting--behavior groups across contrasts and quantify data-group uncertainty conditional on the audited models. Prompt templates, rendering rules, and a worked attribution example appear in Appendix~\ref{sec:prompt-templates}.

\section{Results}

\subsection{Core Audit: Identifier-Linked Effects Emerge at Later Tested Layers and Concentrate in a Low-Rank Subspace}
Across all five alternative A/B/C mappings, CAA produces positive target-versus-source margin shifts at the extraction indices (Table~\ref{tab:letter-permutations}). The corresponding random-adjusted effect under the extraction encoding is 4.63; after remapping, extraction-index effects range from 3.00 to 4.13 and exceed semantic-label effects by 1.95--4.04. This ordering holds for every remapping on the strict split.

Because semantic-label and extraction-index effects coincide under the extraction encoding, only the alternative mappings distinguish them.

\begin{figure*}[!t]
\centering
\includegraphics[width=.97\linewidth]{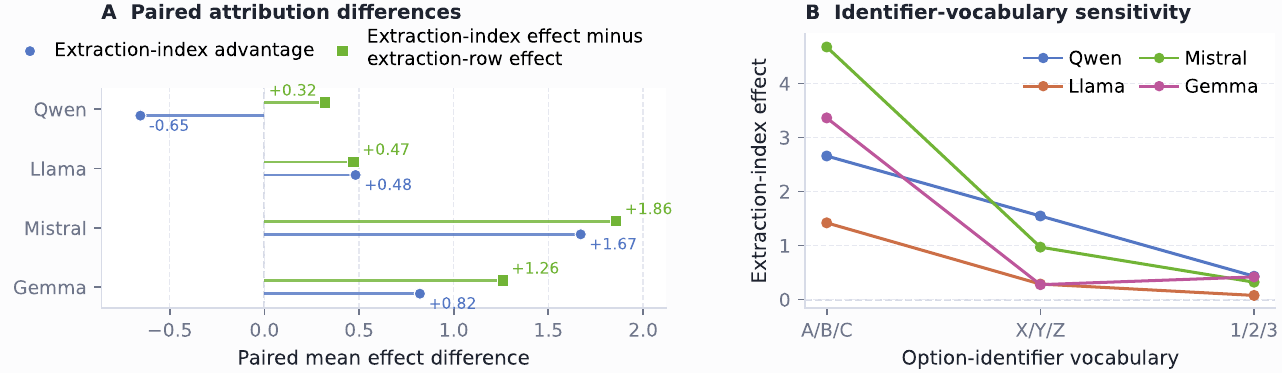}
\caption{NormBank factorial attribution on the mean-token log-likelihood-margin scale. Panel A reports extraction-index advantage (extraction-index effect minus semantic-label effect) and extraction-index effect minus extraction-row effect. Positive values indicate a larger extraction-index effect than semantic-label or extraction-row effect, respectively. Panel B shows the extraction-index effect across option-identifier vocabularies. All Panel A group-cluster intervals and all 12 gated Panel B bootstrap intervals exclude zero; the corresponding Panel A sign-flip tests remain significant after Holm correction.}
\label{fig:factorial-attribution}
\end{figure*}

The factorial audit attributes this pattern to identifier index rather than row position (Figure~\ref{fig:factorial-attribution}): the extraction-index effect exceeds the extraction-row effect in every model by 0.32--1.86. Llama, Mistral, and Gemma also favor the extraction index over semantic-label following, while Qwen favors semantic-label following by 0.65. X/Y/Z and 1/2/3 weaken the extraction-index effect in every model. Before steering, every model exceeds three-way chance under all six A/B/C mappings. 
To check that the ordering is not driven by items on which the unsteered model fails under one or more remappings, we restrict each model to pairs it solves correctly under all six mappings. On this subset (309 unique pairs overall), the pooled extraction-index effect remains 1.455 [1.443, 1.467] and its advantage .426 [.408, .443].

\begin{figure*}[!t]
\centering
\includegraphics[width=.97\linewidth]{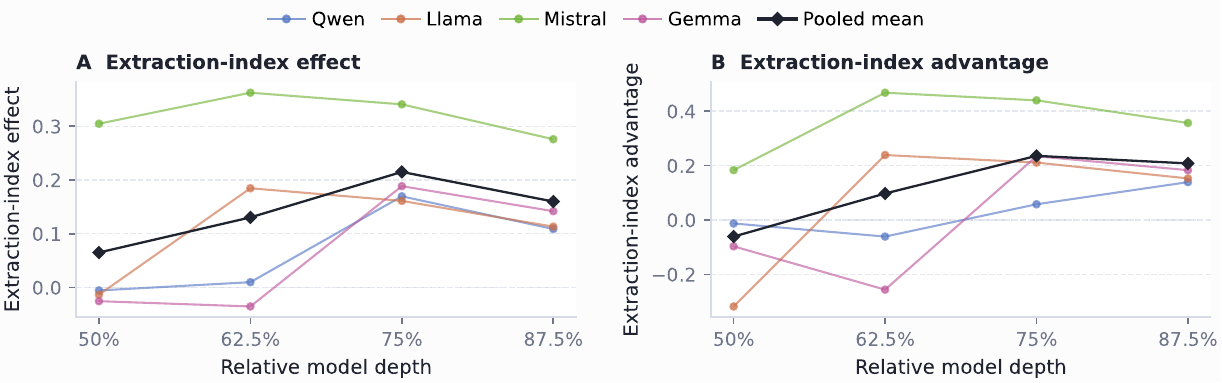}
\caption{Norm-matched layer-wise NormBank attribution on the option-choice probability-margin scale. Each model--contrast direction is rescaled to its 75\%-depth norm. Colored lines show fixed-model trajectories, and black diamonds show pooled means. Panel A reports the extraction-index effect; Panel B reports extraction-index advantage. The corresponding 95\% setting--behavior group-cluster intervals appear in Appendix~\ref{sec:layer-attribution}.}
\label{fig:layer-attribution}
\end{figure*}

\paragraph{Extraction-index following emerges at later layers.}
After norm matching, the extraction-index effect changes across the four depths from 0.065 to 0.130, 0.215, and 0.160 (Figure~\ref{fig:layer-attribution}); its advantage changes from $-0.061$ to $+0.098$, $+0.236$, and $+0.208$. 
Both the extraction-index effect and advantage are larger at 75\% and 87.5\% depth than at 50\%, with paired group-cluster intervals excluding zero.
Transitions differ by model, but all four show positive extraction-index effect and advantage at 75\% and 87.5\%.

\paragraph{Position localization.}
Moving direction extraction from pre-answer to scenario end reduces the extraction-index probability-margin effect from 0.218 to 0.004 (98.3\%). The effect remains near zero after norm rescaling and matched-position injection. The pre-answer extraction-index effect exceeds the scenario-end effect in all 20 model--mapping cells. This localizes the extraction-index profile to late pre-answer states rather than to direction norm or an injection-position mismatch.
At the pre-answer site, the hidden state has already incorporated the question, displayed option identifiers, label descriptions, and output instruction; the scenario-end site precedes this answer scaffold.

\paragraph{Direct output-sensitive baseline.}
On the same option-choice probability-margin scale at 75\% depth, a norm-matched mean local-gradient controller produces a 0.417 extraction-index effect and 0.526 extraction-index advantage, compared with 0.215 and 0.236 for canonical CAA. A simple output-sensitive controller therefore reproduces and amplifies the same extraction-index-over-semantic-label ordering. The mean local-gradient direction is only moderately aligned with full CAA (mean cosine .329) but aligns much more strongly with the CAA readout projection (.845).
This baseline establishes that local output sensitivity is sufficient to reproduce the ordering; the following decomposition tests whether canonical CAA relies on the same geometry.

\paragraph{Effect concentration in the output-sensitive subspace.}
The validation-selected projection contains 15.4\% of the direction's squared $L_2$ energy but retains 96.3\% [95.8, 96.7] of its extraction-index effect; the residual contains 84.6\% of the energy and retains 1.0\% [0.6, 1.3]. 
Most of the direction norm therefore lies outside the selected subspace, but almost none of the measured extraction-index effect does. A fixed rank-2 basis also preserves a clear projection--residual separation.
Its extraction-index effect exceeds that of covariance-matched controls. On validation prompts, first-order Jacobian predictions correlate with observed effects across $\alpha\in\{.2,.4,.8\}$ (mean Pearson $r=.83$--$.96$). 

\paragraph{Cross-vocabulary transfer.}
The A/B/C-derived projection remains stronger than the residual when the identifier vocabulary changes. Among cells passing the mapping-comprehension check under all three vocabularies, it retains 95.0\% of unprojected CAA with X/Y/Z and 83.3\% with 1/2/3, exceeding the residual in every model-vocabulary aggregate. Both full and projected effects attenuate under the new vocabularies. 
Without competence filtering, the projection-over-residual ordering also holds in all 12 model--vocabulary aggregates.

\begin{table}[t]
\centering
\small
\setlength{\tabcolsep}{2pt}
\begin{tabular}{@{}lccc@{}}
\toprule
Dataset
& Pair regime
& \begin{tabular}[t]{@{}c@{}}
    Semantic-label\\
    $>0$
  \end{tabular}
& \begin{tabular}[t]{@{}c@{}}
    Extraction-index\\
    advantage $>0$
  \end{tabular} \\
\midrule
NormBank & Strict context      & 3/5 & 5/5 \\
\begin{tabular}[c]{@{}l@{}}
  NormBank\\
  (ITI)
\end{tabular}
& Strict context & 2/5 & 5/5 \\
MNLI     & Same-premise pairs & 3/5 & 5/5 \\
SC101    & Action-level pairs  & 5/5 & 0/5 \\
\bottomrule
\end{tabular}
\caption{Random-adjusted patterns over five alternative mappings. Rows use CAA unless marked ITI. Counts are mappings with positive point estimates, not significance decisions. ITI averages the three models retained on validation.}
\label{tab:cross-dataset-profiles}
\end{table}

ITI reproduces the NormBank ordering in all three retained models: extraction-index advantage is positive under all five alternative mappings, whereas the semantic-label effect is positive in only two of the five mappings. On MNLI same-premise pairs, pooled semantic-label effects are positive in 3/5 mappings and extraction-index advantage in 5/5; all 24 unsteered model--mapping accuracy intervals exceed chance. Gemma, Llama, and Mistral favor the extraction index under all five mappings, whereas Qwen favors semantic-label following. 
SC101 reverses the aggregate ordering (Table~\ref{tab:cross-dataset-profiles}), confirming that attribution varies across evaluation regimes.

\subsection{Behavioral Validation: MCQ and Open-Ended Verdicts Can Diverge in a Public CAA Protocol}
The published CAA directions receive different verdicts from multiple-choice and open-ended evaluations (Table~\ref{tab:published-caa-summary}). After equal weighting across the two models, the three-judge median effect is positive for hallucination ($+1.60$ [0.84, 2.44]) and sycophancy ($+1.11$ [0.57, 1.75]), but not refusal ($-0.13$ [$-0.54$, 0.18]). Hallucination is positive in both evaluations. Refusal has a large original A/B gain without an open-ended increase, whereas sycophancy is non-positive under the original metric but increases in open generation. On 180 blinded responses, the judge panel agreed closely with adjudicated human scores (Pearson $r=.927$; quadratic-weighted $\kappa=.929$).
Thus, multiple-choice steering effects need not imply the same behavioral conclusion under open-ended generation. Judge and human-validation details appear in Appendix~\ref{sec:supp-behavioral}.

\begin{table}[t]
\centering
\small
\setlength{\tabcolsep}{3pt}
\begin{tabular}{@{}lrr@{}}
\toprule
Behavior & Original MCQ & Open-ended [95\% CI] \\
\midrule
Hallucination & $+3.31$ & $+1.60$ [$+0.84$, $+2.44$] \\
Refusal & $+3.28$ & $-0.13$ [$-0.54$, $+0.18$] \\
Sycophancy & $-0.48$ & $+1.11$ [$+0.57$, $+1.75$] \\
\bottomrule
\end{tabular}
\caption{Public CAA verdicts over two Llama-2 models. Original MCQ is the $+1$-dose change in the published A/B mean-token log-likelihood margin. Open-ended effects use three judges validated on 180 blinded human-scored responses. Columns retain protocol-specific scales and doses; we compare their qualitative verdicts.}
\label{tab:published-caa-summary}
\end{table}

\section{Discussion}
On NormBank, extraction-index following is weak before the answer scaffold and concentrates at late pre-answer states in directions locally sensitive to identifier logits. A direct gradient baseline reproduces the ordering, and the selected subspace carries nearly all of the effect; removing it does not increase semantic-label following. This functionally localizes the dependence without identifying a complete circuit. MNLI extends the pooled pattern, while Qwen and SC101 show model- and evaluation-regime-dependent attribution.

The audit separates two questions that are conflated when extraction and evaluation share an answer encoding. A direction may produce a strong score change under that encoding, yet the change may be carried by an output-sensitive identifier component rather than consistently producing semantic-label following. Extraction-index following is therefore an evaluation-level attribution, not evidence that the model represents an abstract index variable.
The contrasting MNLI, SC101, and Qwen profiles show that this attribution must be measured rather than assumed. More broadly, the evidence required should match the level of control being claimed (Table~\ref{tab:evaluation-protocol}).

\begin{table}[t]
\centering
\small
\setlength{\tabcolsep}{3pt}
\begin{tabularx}{\columnwidth}{@{}lX@{}}
\toprule
Claim & Recommended evidence \\
\midrule
Target-score movement & Target-margin effect under the extraction encoding against $L_2$-matched random controls \\
Answer-encoding attribution & Frozen remapping; factorial identifier/row controls; position and subspace localization \\
Selective control & Context-matched tests of selective response differences \\
Generative control & Open generations evaluated independently of multiple-choice steering scores \\
\bottomrule
\end{tabularx}
\caption{Claim--evidence matching for activation-steering evaluation.}
\label{tab:evaluation-protocol}
\end{table}

\section{Limitations}
The six-mapping audit covers three three-label tasks and four instruction-tuned model families; the full factorial audit, localization analyses, and ITI use NormBank. The gradient-defined subspace functionally localizes the effect relative to the identifier readout at the tested layers and positions, without identifying a complete circuit. Layer-wise norm matching equates direction magnitude but not layer-specific model sensitivity, and position localization uses a single-site intervention. The public CAA comparison follows protocol-specific doses.

\section{Conclusion}
Cross-encoding evaluation freezes an intervention and re-encodes answers to the same held-out items, separating semantic-label following from extraction-index following. On NormBank, CAA preferentially follows extraction indices, an ordering replicated by ITI in three models; the effect emerges at later tested depths and concentrates in a low-rank output-sensitive component. MNLI reproduces the pooled ordering despite a Qwen reversal, while SC101 favors semantic-label following, showing that attribution varies across models and tasks. Open-generation verdicts can also diverge from multiple-choice scores. Steering claims therefore require evidence that directly tests the claimed form of control.

\clearpage
\bibliographystyle{plainnat}
\bibliography{references}

@inproceedings{panickssery2024caa,
  title={Steering Llama 2 via Contrastive Activation Addition},
  author={Rimsky, Nina and Gabrieli, Nick and Schulz, Julian and Tong, Meg and Hubinger, Evan and Turner, Alexander},
  booktitle={Proceedings of the 62nd Annual Meeting of the Association for Computational Linguistics (Volume 1: Long Papers)},
  year={2024},
  address={Bangkok, Thailand},
  publisher={Association for Computational Linguistics},
  pages={15504--15522},
  doi={10.18653/v1/2024.acl-long.828},
  url={https://aclanthology.org/2024.acl-long.828/}
}

@inproceedings{xu2026steereval,
  title={How Controllable Are Large Language Models? A Unified Evaluation across Behavioral Granularities},
  author={Xu, Ziwen and Xu, Kewei and Xu, Haoming and Hong, Haiwen and Huang, Longtao and Xue, Hui and Zhang, Ningyu and Shen, Yongliang and Zheng, Guozhou and Chen, Huajun and Deng, Shumin},
  booktitle={Proceedings of the 64th Annual Meeting of the Association for Computational Linguistics (Volume 1: Long Papers)},
  year={2026},
  address={San Diego, California, United States},
  publisher={Association for Computational Linguistics},
  pages={31269--31299},
  doi={10.18653/v1/2026.acl-long.1443},
  url={https://aclanthology.org/2026.acl-long.1443/}
}

@inproceedings{li2024verbalizer,
  title={Instruction-following Evaluation through Verbalizer Manipulation},
  author={Li, Shiyang and Yan, Jun and Wang, Hai and Tang, Zheng and Ren, Xiang and Srinivasan, Vijay and Jin, Hongxia},
  booktitle={Findings of the Association for Computational Linguistics: NAACL 2024},
  year={2024},
  address={Mexico City, Mexico},
  publisher={Association for Computational Linguistics},
  pages={3678--3692},
  doi={10.18653/v1/2024.findings-naacl.233},
  url={https://aclanthology.org/2024.findings-naacl.233/}
}

@article{zou2023representation,
  title={Representation Engineering: A Top-Down Approach to AI Transparency},
  author={Zou, Andy and Phan, Long and Chen, Sarah and Campbell, James and Guo, Phillip and Ren, Richard and Pan, Alexander and Yin, Xuwang and Mazeika, Mantas and Dombrowski, Ann-Kathrin and Goel, Shashwat and Li, Nathaniel and Byun, Michael J. and Wang, Zifan and Mallen, Alex and Basart, Steven and Koyejo, Sanmi and Song, Dawn and Fredrikson, Matt and Kolter, J. Zico and Hendrycks, Dan},
  journal={arXiv preprint arXiv:2310.01405},
  year={2023},
  url={https://arxiv.org/abs/2310.01405}
}

@article{tan2024steeringreliability,
  title={Analyzing the Generalization and Reliability of Steering Vectors},
  author={Tan, Daniel and Chanin, David and Lynch, Aengus and Kanoulas, Dimitrios and Paige, Brooks and Garriga-Alonso, Adria and Kirk, Robert},
  journal={arXiv preprint arXiv:2407.12404},
  year={2024},
  url={https://arxiv.org/abs/2407.12404}
}

@article{zheng2023mcqselectors,
  title={Large Language Models Are Not Robust Multiple Choice Selectors},
  author={Zheng, Chujie and Zhou, Hao and Meng, Fandong and Zhou, Jie and Huang, Minlie},
  journal={arXiv preprint arXiv:2309.03882},
  year={2023},
  url={https://arxiv.org/abs/2309.03882}
}

@article{dang2026cultural,
  title={Cultural Value Alignment Via Latent Activation Steering in Large Language Models},
  author={Dang, Trung Duc Anh and Masud, Sarah},
  journal={arXiv preprint arXiv:2605.26365},
  year={2026},
  url={https://arxiv.org/abs/2605.26365}
}

@article{dang2026scenario,
  title={Scenario-based Probing and Steering Cultural Values in Large Language Models--Extended Version},
  author={Dang, Trung Duc Anh and Kieu, Tung and Masud, Sarah},
  journal={arXiv preprint arXiv:2606.11399},
  year={2026},
  url={https://arxiv.org/abs/2606.11399}
}

@article{tlaie2024moralcompass,
  title={Exploring and Steering the Moral Compass of Large Language Models},
  author={Tlaie, Alejandro},
  journal={arXiv preprint arXiv:2405.17345},
  year={2024},
  url={https://arxiv.org/abs/2405.17345}
}

@article{sauter2026rules,
  title={Between Rules and Reality: On the Context Sensitivity of LLM Moral Judgment},
  author={Sauter, Adrian and Schirmer, Mona},
  journal={arXiv preprint arXiv:2603.23114},
  year={2026},
  url={https://arxiv.org/abs/2603.23114}
}

@article{jin2025conva,
  title={Internal Value Alignment in Large Language Models through Controlled Value Vector Activation},
  author={Jin, Haoran and Li, Meng and Wang, Xiting and Xu, Zhihao and Huang, Minlie and Jia, Yantao and Lian, Defu},
  journal={arXiv preprint arXiv:2507.11316},
  year={2025},
  url={https://arxiv.org/abs/2507.11316}
}

@article{yang2026neva,
  title={Controllable Value Alignment in Large Language Models through Neuron-Level Editing},
  author={Yang, Yonghui and Wang, Yihui and Li, Junwei and Liu, Jilong and Zhu, Fengbin and Huang, Weibiao and Wu, Le and Hong, Richang and Chua, Tat-Seng},
  journal={arXiv preprint arXiv:2602.07356},
  year={2026},
  url={https://arxiv.org/abs/2602.07356}
}

@article{belrose2023tunedlens,
  title={Eliciting Latent Predictions from Transformers with the Tuned Lens},
  author={Belrose, Nora and Furman, Zach and Smith, Logan and Halawi, Danny and Ostrovsky, Igor and McKinney, Lev and Biderman, Stella and Steinhardt, Jacob},
  journal={arXiv preprint arXiv:2303.08112},
  year={2023},
  url={https://arxiv.org/abs/2303.08112}
}

@article{park2023linear,
  title={The Linear Representation Hypothesis and the Geometry of Large Language Models},
  author={Park, Kiho and Choe, Yo Joong and Veitch, Victor},
  journal={arXiv preprint arXiv:2311.03658},
  year={2023},
  url={https://arxiv.org/abs/2311.03658}
}

@article{kramar2024atp,
  title={{AtP*}: An Efficient and Scalable Method for Localizing {LLM} Behaviour to Components},
  author={Kram{\'a}r, J{\'a}nos and Lieberum, Tom and Shah, Rohin and Nanda, Neel},
  journal={arXiv preprint arXiv:2403.00745},
  year={2024},
  url={https://arxiv.org/abs/2403.00745}
}

@article{ziems2023normbank,
  title={NormBank: A Knowledge Bank of Situational Social Norms},
  author={Ziems, Caleb and Dwivedi-Yu, Jane and Wang, Yi-Chia and Halevy, Alon and Yang, Diyi},
  journal={arXiv preprint arXiv:2305.17008},
  year={2023},
  url={https://arxiv.org/abs/2305.17008}
}

@inproceedings{forbes2020socialchemistry,
  title={Social Chemistry 101: Learning to Reason about Social and Moral Norms},
  author={Forbes, Maxwell and Hwang, Jena D. and Shwartz, Vered and Sap, Maarten and Choi, Yejin},
  booktitle={Proceedings of the 2020 Conference on Empirical Methods in Natural Language Processing},
  year={2020},
  url={https://arxiv.org/abs/2011.00620}
}

@article{qwen2024qwen25,
  title={Qwen2.5 Technical Report},
  author={{Qwen Team}},
  journal={arXiv preprint arXiv:2412.15115},
  year={2024},
  url={https://arxiv.org/abs/2412.15115}
}

@article{dubey2024llama3,
  title={The Llama 3 Herd of Models},
  author={Dubey, Abhimanyu and others},
  journal={arXiv preprint arXiv:2407.21783},
  year={2024},
  url={https://arxiv.org/abs/2407.21783}
}

@article{jiang2023mistral,
  title={Mistral 7B},
  author={Jiang, Albert Q. and Sablayrolles, Alexandre and Mensch, Arthur and others},
  journal={arXiv preprint arXiv:2310.06825},
  year={2023},
  url={https://arxiv.org/abs/2310.06825}
}

@article{gemma2024gemma2,
  title={Gemma 2: Improving Open Language Models at a Practical Size},
  author={{Gemma Team}},
  journal={arXiv preprint arXiv:2408.00118},
  year={2024},
  url={https://arxiv.org/abs/2408.00118}
}

@inproceedings{li2023iti,
  title={Inference-Time Intervention: Eliciting Truthful Answers from a Language Model},
  author={Li, Kenneth and Patel, Oam and Vi{\'e}gas, Fernanda and Pfister, Hanspeter and Wattenberg, Martin},
  booktitle={Advances in Neural Information Processing Systems},
  volume={36},
  year={2023},
  url={https://arxiv.org/abs/2306.03341}
}

@article{shao2026decodeshare,
  title   = {DecodeShare: Tracing the Shared Subspace of LLM Decode-Time Decisions},
  author  = {Shao, Zishan and Zhang, Lixun and Cui, Kangning and Wang, Yixiao
             and Jiang, Ting and Ye, Hancheng and Wang, Qinsi and Du, Zhixu
             and Fu, Yuzhe and Yang, Fan and Zhuo, Danyang and Chen, Yiran
             and Li, Hai Helen},
  journal = {arXiv preprint arXiv:2607.20469},
  year    = {2026}
}

@article{pres2024reliable,
  title={Towards Reliable Evaluation of Behavior Steering Interventions in LLMs},
  author={Pres, Itamar and Ruis, Laura and Lubana, Ekdeep Singh and Krueger, David},
  journal={arXiv preprint arXiv:2410.17245},
  year={2024}
}

@article{venkatesh2026nonidentifiability,
  title={On the Non-Identifiability of Steering Vectors in Large Language Models},
  author={Venkatesh, Sohan and Kurapath, Ashish Mahendran},
  journal={arXiv preprint arXiv:2602.06801},
  year={2026}
}

@article{fraile2026internal,
  title={Internal Representation, Not Clinical Knowledge: Where Apparent LLM Triage Failures Originate},
  author={Fraile Navarro, David and Como, Berardino and Sheng, Jialei and Ananthan, Soundariya and Berkovsky, Shlomo},
  journal={arXiv preprint arXiv:2605.29889},
  year={2026}
}

@inproceedings{williams2018broad,
  title     = {A Broad-Coverage Challenge Corpus for Sentence Understanding through Inference},
  author    = {Williams, Adina and Nangia, Nikita and Bowman, Samuel R.},
  booktitle = {Proceedings of the 2018 Conference of the North American Chapter of the Association for Computational Linguistics: Human Language Technologies},
  pages     = {1112--1122},
  year      = {2018}
}

@article{ye2026steering,
  title={Where Steering Signals Come From: Activation Source Selection in Activation Steering},
  author={Ye, Jiaran and Ran, Lingxu and Yao, Zijun and Wang, Chenpeng and Jiang, Yong and Hou, Lei and Li, Juanzi and Pan, Liangming},
  journal={arXiv preprint arXiv:2607.25270},
  year={2026}
}

\clearpage
\appendix
\section{Reproducibility and Data Isolation}
\label{sec:supp-reproducibility}

\subsection{Notation Summary}
Table~\ref{tab:notation} collects the quantities used throughout the evaluation. NormBank and MNLI cross-encoding tests and the NormBank factorial audit use mean-token log-likelihood margins. Layer, position, and subspace analyses use normalized option-choice probability margins; SC101 uses its protocol-specific probability scale. The tables and captions state the applicable scale; values are compared only within a scale and answer encoding.

\subsection{Pair Construction and Split Details}
\label{sec:pair-construction-details}
We clean the \texttt{setting}, \texttt{behavior}, \texttt{constraints}, and label fields; retain only \textit{taboo} (T), \textit{normal} (N), and \textit{expected} (E); and group rows by exact \texttt{setting + behavior}. We use the ordered contrasts T$\rightarrow$N, T$\rightarrow$E, and N$\rightarrow$E. Within each sorted group--contrast cell, lower- and higher-label rows are deterministically permuted with fixed seeds and matched one-to-one without replacement. An endpoint is therefore not reused within the same group--contrast matching; the accompanying code records the exact seeds.

The central exhaustive mapping audit assigns complete setting--behavior groups to train, validation, or test and removes cross-split endpoint reuse. Pair IDs, context groups, and endpoints are therefore disjoint across splits (Table~\ref{tab:strict-split-isolation}). The CAA audit uses up to 2,048 training and 512 test pairs per contrast; ITI uses up to 1,024/256/512 train/validation/test pairs. The local identifier-readout analysis uses at most 64 training prompts, 32 validation prompts, and 128 test pairs per contrast. Test prompts never enter direction extraction, basis estimation, rank selection, or normalization.

\begin{table}[!tb]
\centering
\small
\setlength{\tabcolsep}{2.5pt}
\begin{tabularx}{\columnwidth}{@{}lX@{}}
\toprule
Symbol & Meaning \\
\midrule
$\vraw^{(c)}$ & Canonical target-minus-source CAA direction for contrast $c$ \\
$f,T_f(y)$ & Answer encoding and the token sequence expressing semantic label $y$ \\
$m_f(x)$ & Target-versus-source mean log-likelihood margin \\
$\Delta m_f$ & Steered-minus-base margin change \\
$\delta_{s,i,k}^{(v,f)}$ & Pair effect after subtracting random draw $k$ \\
$D_{s,k}^{(v,f)}$ & Standardized detectability within one stratum \\
$D_f^{\mathrm{ctrl}}(v)$ & Within-encoding detectability relative to matched random controls \\
Extraction index & Index in the ordered answer-identifier vocabulary assigned under the extraction encoding; aligned as A/X/1, B/Y/2, C/Z/3 \\
$\Delta m_{\mathrm{sem}}$ & Semantic-label log-likelihood-margin effect \\
$\Delta m_{\mathrm{idx}}$ & Extraction-index log-likelihood-margin effect \\
$A_{\mathrm{idx}}$ & Extraction-index advantage: extraction-index minus semantic-label effect on the stated scale \\
Factorial effects & Semantic-label, extraction-index, and extraction-row mean log-likelihood-margin changes \\
$\mathbf U_r$ & Rank-$r$ basis for the local identifier-readout subspace \\
$\mathbf v_{\parallel}$ & Projection of $\vraw$ into $\operatorname{span}(\mathbf U_r)$ \\
$\mathbf v_{\perp}$ & Component of $\vraw$ orthogonal to $\mathbf U_r$ \\
$E(\mathbf u)$ & Squared $L_2$ energy fraction $\|\mathbf u\|_2^2/\|\vraw\|_2^2$ \\
JS shift & Jensen--Shannon divergence between normalized option-choice distributions before and after steering (natural logarithms) \\
$G_f$ & Change in higher-versus-lower context separation \\
$\Delta G_f$ & Random-adjusted separation change relative to Gaussian controls \\
$\Delta A_f$ & Random-adjusted pair-ranking accuracy change in percentage points \\
\bottomrule
\end{tabularx}
\caption{Notation used in the main paper and appendix controls.}
\label{tab:notation}
\end{table}

The factorial attribution audit reuses the strict CAA directions and the same 1,536 test pairs. It crosses six label-to-identifier mappings, three identifier vocabularies (A/B/C, X/Y/Z, and 1/2/3), and all six displayed row orders, producing 108 conditions for each frozen direction. Canonical direction extraction always uses A/B/C=\textit{taboo}/\textit{normal}/\textit{expected}; no factorial test condition reconstructs or retunes it. The constructive control forms a mapping-balanced direction by extracting under all six mappings on the strict training pairs, averaging and norm-rescaling the directions, and rerunning the same 108 test conditions.

The direct-label, opaque-codeword, and matched-context analyses use a separate pair-ID split. That split supports these analyses, while the strict group-disjoint split isolates the exhaustive letter-mapping and mapping-balanced factorial tests at the contextual-content level. Four of the 1,536 pair-ID-split test pairs have identical rendered endpoint text despite conflicting labels in the public source. They remain valid labeled items for target-score movement but cannot define a discrimination pair, so the matched-context diagnostic uses 1,532 pairs. Every answer encoding evaluates the same test pair IDs with three prompt templates. The published CAA audit uses all available open-ended items and up to 50 multiple-choice items per behavior.

MNLI uses the three relation labels \textit{entailment}, \textit{neutral}, and \textit{contradiction}. Within each normalized premise group, we pair hypotheses carrying different labels and assign the complete premise group to one split, so no premise crosses train and test. Each contrast uses up to 2,048 training and 512 test pairs. The direction is extracted once under A/B/C=\textit{entailment}/\textit{neutral}/\textit{contradiction} and evaluated under all five alternative mappings with the same four models, layers, dose, three prompt templates, and five norm-matched random controls used by the primary CAA audit. Intervals resample complete premise groups and weight model--contrast strata equally.

The SC101 scope replication uses action-only statements labeled \textit{bad}, \textit{ok}, or \textit{good}. It forms the three pairwise label contrasts and selects up to 2,048 training and 256 test pairs per contrast. These pairs do not hold context fixed; they test whether a frozen direction exhibits semantic-label or extraction-index following under all six label-to-identifier mappings. Matched-context selectivity is evaluated on NormBank, whose pairs hold setting and behavior fixed. 

\begin{table}[!htbp]
\centering
\small
\setlength{\tabcolsep}{2pt}
\begin{tabularx}{\columnwidth}{@{}lXX@{}}
\toprule
Setting / behavior & Lower endpoint & Higher endpoint \\
\midrule
Boat / Cook food & \texttt{role = engineer} (taboo) & \texttt{role = chef} (expected) \\
\bottomrule
\end{tabularx}
\caption{NormBank matched-pair example; only the contextual constraint and label differ.}
\label{tab:normbank-pair-example}
\end{table}

\begin{table}[!htbp]
\centering
\small
\setlength{\tabcolsep}{3pt}
\begin{tabular}{@{}lrrr@{}}
\toprule
Split & Pairs & Context groups & Endpoints \\
\midrule
Train & 75,868 & 10,620 & 105,487 \\
Validation & 9,796 & 1,328 & 13,622 \\
Test & 9,353 & 1,302 & 12,799 \\
\midrule
Cross-split overlap & 0 & 0 & 0 \\
\bottomrule
\end{tabular}
\caption{Strict NormBank split inventory before experiment-specific sampling. The final row audits overlap in pair IDs, setting--behavior groups, and endpoints; all three isolation checks pass.}
\label{tab:strict-split-isolation}
\end{table}

\subsection{Locked Model Settings}
\label{sec:locked-settings}
All CAA model-specific choices are fixed before evaluation under any test answer encoding. The primary audit uses the depth-matched layer nearest 75\% of the decoder stack and the shared strength $\alpha=0.8$; contrast orientation determines direction sign. Every strict-split contrast uses up to 2,048 training pairs, 512 test pairs, and three prompt templates.

\begin{table}[!htbp]
\centering
\small
\begin{tabular}{@{}lcc@{}}
\toprule
Model & Layer & $\alpha$ \\
\midrule
Qwen2.5-7B & 20 & 0.8 \\
Llama-3.1-8B & 23 & 0.8 \\
Mistral-7B-v0.3 & 23 & 0.8 \\
Gemma-2-9B & 31 & 0.8 \\
\bottomrule
\end{tabular}
\caption{Prespecified depth-matched settings for the controlled audit.}
\label{tab:locked-settings}
\end{table}

MNLI reuses the primary settings in Table~\ref{tab:locked-settings}. The SC101 scope replication instead freezes Qwen layer 20, Llama layer 16, Mistral layer 16, and Gemma layer 20 with $\alpha=1.0$. These middle-layer settings were fixed before the five alternative mappings were evaluated. They differ from the NormBank and MNLI settings, so SC101 contributes evidence about the label-versus-identifier ordering and its sign consistency, not a cross-dataset magnitude comparison.

\subsection{Prompt Templates and Answer Rendering}
\label{sec:prompt-templates}
The controlled audits render prompts deterministically from the templates below. Braced fields are replaced by the corresponding dataset text. Candidate answers are not included in the prompt after \texttt{Answer:}; they are scored as alternative continuations of that same prefix. Changing an answer encoding therefore changes how labels are expressed, while the scoring rule remains sequence likelihood under the displayed encoding.

\begin{table*}[!t]
\centering
\small
\setlength{\tabcolsep}{4pt}
\begin{tabularx}{\textwidth}{@{}p{0.15\textwidth}X@{}}
\toprule
Dataset / use & Rendered prompt skeleton \\
\midrule
NormBank direction extraction &
\texttt{You are classifying a behavior under a social-norm context.}\newline
\texttt{Scenario: \{scenario\}}\newline
\texttt{Question: In this context, how should the behavior be classified?}\newline
\texttt{A. taboo - socially forbidden or unacceptable}\newline
\texttt{B. normal - socially ordinary or acceptable}\newline
\texttt{C. expected - socially expected or encouraged}\newline
\texttt{Answer with only A, B, or C.}\newline \texttt{Answer:} \\
\addlinespace
NormBank factorial audit &
\texttt{You are classifying a behavior under a social-norm context.}\newline
\texttt{Scenario: \{scenario\}}\newline
\texttt{Question: In this context, how should the behavior be classified?}\newline
\texttt{\{identifier in displayed row 1\}. \{label description\}}\newline
\texttt{\{identifier in displayed row 2\}. \{label description\}}\newline
\texttt{\{identifier in displayed row 3\}. \{label description\}}\newline
\texttt{Use the answer key shown below.}\newline
\texttt{Return exactly one identifier from: \{identifier set\}.}\newline \texttt{Answer:} \\
\addlinespace
MNLI direction extraction &
\texttt{Premise: \{premise\}}\newline
\texttt{Hypothesis: \{hypothesis\}}\newline
\texttt{Question: What is the logical relationship between the premise and hypothesis?}\newline
\texttt{A. entailment - the hypothesis follows from the premise}\newline
\texttt{B. neutral - neither entailment nor contradiction}\newline
\texttt{C. contradiction - the hypothesis conflicts with the premise}\newline
\texttt{Return exactly one answer.}\newline \texttt{Answer:} \\
\addlinespace
SC101 direction extraction &
\texttt{You are judging a social norm.}\newline
\texttt{Action: \{action\}}\newline
\texttt{Question: Is this action bad, ok, or good?}\newline
\texttt{A. bad - socially inappropriate or harmful}\newline
\texttt{B. ok - socially neutral or acceptable}\newline
\texttt{C. good - socially appropriate or beneficial}\newline
\texttt{Answer with only A, B, or C.}\newline \texttt{Answer:} \\
\bottomrule
\end{tabularx}
\caption{Exact extraction skeletons and the factorial evaluation skeleton. Line breaks match the rendered prompts. Alternative semantic mappings replace the option assignment while preserving all other text.}
\label{tab:prompt-skeletons}
\end{table*}

The main NormBank and MNLI mapping audits evaluate each item with three fixed question--instruction pairs. NormBank uses: (i) ``Question: In this context, how should the behavior be classified?'' followed by ``Return exactly one answer.''; (ii) ``Question: What is this behavior's social-norm status in the stated context?'' followed by ``Provide one classification only.''; and (iii) ``Question: Choose the norm label that best fits the behavior in context.'' followed by ``Give exactly one label.'' MNLI uses: (i) the relationship question shown in Table~\ref{tab:prompt-skeletons}; (ii) ``Question: Classify whether the premise entails, contradicts, or is neutral toward the hypothesis.'' followed by ``Provide one classification only.''; and (iii) ``Question: Choose the inference label that best describes this pair.'' followed by ``Give exactly one label.'' The surrounding context, answer encoding, and final \texttt{Answer:} prefix remain unchanged across these wording variants.

For the six-way letter audits, identifier indices 1--3 receive the semantic labels in the orders T/N/E, T/E/N, N/T/E, N/E/T, E/T/N, and E/N/T for NormBank; MNLI and SC101 use the analogous six permutations of their three labels. The factorial audit independently crosses those mappings with A/B/C, X/Y/Z, and 1/2/3, using the index correspondence A/X/1, B/Y/2, and C/Z/3, and with all six displayed row orders. Its scenario-free competence check asks, for each label, ``Which identifier means \{label\}?'', ``Return the identifier assigned to \{label\}.'', and ``If the correct category is \{label\}, what should be returned?''

\paragraph{Worked attribution example.}


Consider taboo$\rightarrow$expected steering. Extraction uses A/B/C = taboo/normal/expected, with the options displayed in the same order:

\[
\begin{array}{c@{\quad}c@{\quad}c}
\textbf{Index} & \textbf{Ident.} & \textbf{Label} \\
\hline
1 & A & \text{taboo} \\
2 & B & \text{normal} \\
3 & C & \text{expected}
\end{array}
\]

\noindent
Thus the source/target identifiers A/C, extraction indices 1/3, and displayed rows 1/3 coincide.

\[
\Downarrow\quad
\begin{gathered}
\text{freeze direction;}\\
\text{change mapping, vocabulary, and rows}
\end{gathered}
\]

\noindent
Now use X/Y/Z with X = expected, Y = taboo, and Z = normal, and display the options as Z, X, Y:

\[
\begin{array}{c@{\quad}c@{\quad}c@{\quad}c}
\textbf{Row} & \textbf{Ident.} & \textbf{Index} & \textbf{Label} \\
\hline
1 & Z & 3 & \text{normal} \\
2 & X & 1 & \text{expected} \\
3 & Y & 2 & \text{taboo}
\end{array}
\]

\noindent
Let $s(o)$ denote the mean-token log likelihood of candidate identifier $o$. The same taboo$\rightarrow$expected contrast now gives
\[
\begin{array}{@{}ll@{}}
\text{Semantic label:}
  & s(X)-s(Y),\\
\text{Extraction index:}
  & s(Z)-s(X),\\
\text{Extraction row:}
  & s(Y)-s(Z).
\end{array}
\]

\noindent
The three attribution targets are therefore distinct: the current semantic target is X, extraction index 3 is instantiated by Z, and extraction-time row 3 now contains Y. The 1/2/3 vocabulary is analogous: 1, 2, and 3 are answer identifiers, not displayed row numbers, with index alignment A/X/1, B/Y/2, C/Z/3.

For completion encodings, the NormBank prompt replaces the option list with ``Valid completions: \{candidates\}.'' The rendered candidate order is \textit{expected, normal, taboo} for direct labels; \textit{wug, blicket, dax} after the key ``dax means taboo; blicket means normal; wug means expected''; and \textit{3, 2, 1} after defining 1, 2, and 3 as taboo, normal, and expected. Candidate ordering does not affect the separately scored continuations. Each sequence is scored after the same final \texttt{Answer:} prefix.

\paragraph{Tokenization, scoring, and hook semantics.}
The controlled audits tokenize the rendered strings directly with each model's \texttt{AutoTokenizer}; they do not apply an additional chat template. This preserves the same answer scaffold across model families and avoids adding model-specific wrapper tokens to the encoding manipulation. The resulting evidence therefore concerns this controlled prompting protocol. Special tokens are enabled, and a missing padding token is set to the tokenizer's EOS token. A scored sequence is formed by exact string concatenation, \texttt{full\_text = prefix + candidate}, with no implicit separator inserted after \texttt{Answer:}. We verify that tokenizing the concatenation preserves the complete tokenized prefix. Single-token candidates are scored from the next-token distribution; multi-token candidates use the sum of the autoregressive token log likelihoods divided by the number of candidate tokens. Thus, no multi-token candidate is approximated by its final token.

Layer numbers are zero-based decoder-block indices. In the primary controlled audit, the extracted state is the block-output residual stream at the final active token of the prefix ending in \texttt{Answer:}. At evaluation, the forward hook adds $\alpha\mathbf v$ once at that same prefix position. It does not add the direction to later candidate tokens. Padding-aware active-token indices are derived from the attention mask; position-control runs instead resolve the configured scenario-end character boundary to its token position.

The published CAA replication preserves each released source question. The original and swapped multiple-choice conditions retain the two answer texts and either preserve or exchange their A/B identifiers. The direct-answer condition appends ``Choose exactly one of these complete answers:'', followed by the two answer texts, and ``Respond with the complete chosen answer.'' The opaque condition appends ``Temporary answer key:'', defines \textit{dax} and \textit{blicket} as the two complete answers, and requests exactly one codeword. Open-ended generation uses the released question without an answer list. All Llama-2 inputs use the same \texttt{[INST] ... [/INST]} wrapper as the published implementation. Following that protocol, the intervention is active continuously from the assistant-response boundary through completion scoring or generation. Open responses use greedy decoding with at most 100 new tokens and EOS stopping; interface scoring rejects sequences exceeding 2,048 tokens. Each automated judge receives the behavior rubric, user question, and assistant response in that order and returns one JSON object with a short \texttt{reason} followed by an integer \texttt{score} from 0 to 10.

\subsection{Answer-Key Competence Checks}
Opaque completions are interpretable only when the unsteered model can decode the temporary key. The NormBank test asks each model to identify all three codeword meanings under both evaluated mappings; all eight model--mapping cells pass the exact-comprehension criterion (Table~\ref{tab:opaque-competence}). The published CAA models fail a stricter counterbalanced key test (Table~\ref{tab:caa-opaque-competence}), so their opaque results are excluded from the open-ended behavioral evidence.

\begin{table}[!tb]
\centering
\small
\setlength{\tabcolsep}{2pt}
\begin{tabularx}{\columnwidth}{@{}lcccX@{}}
\toprule
Dataset & Conditions & Cells & Pass & Use \\
\midrule
NormBank & 2 mappings & 8 & 8/8 & Retained \\
Published CAA & 4 conditions & 2 & 0/2 & Sensitivity only \\
\bottomrule
\end{tabularx}
\caption{Opaque-key competence summary; strict pass requires correct decoding for every tested target and key.}
\label{tab:opaque-competence}
\end{table}

\begin{table}[!htbp]
\centering
\small
\setlength{\tabcolsep}{2pt}
\begin{tabular}{@{}lcccc@{}}
\toprule
Model & Acc. [95\% CI] & A/B & Fwd./Rev. & Pass \\
\midrule
13B & .712 [.697, .725] & .423/1.000 & .923/.500 & No \\
7B & .747 [.737, .755] & .510/.983 & .990/.503 & No \\
\bottomrule
\end{tabular}
\caption{Published CAA opaque-key competence for Llama-2 over 600 counterbalanced queries per model.}
\label{tab:caa-opaque-competence}
\end{table}

\subsection{Computing Environment}
\label{sec:computing-environment}
Experiments were run on a server with two NVIDIA A100-PCIE-40GB GPUs (40\,GB each), an AMD EPYC 7702P 64-Core CPU, and 503\,GB RAM. The software environment was Ubuntu 22.04.5 LTS, CUDA 12.2, Python 3.11.7, PyTorch 2.7.0, and Transformers 4.52.4. GPU visibility was selected externally with \texttt{CUDA\_VISIBLE\_DEVICES}; the accompanying scripts do not assume a fixed device index. The controlled audit scores each frozen model under three fixed prompt templates; uncertainty is obtained by resampling held-out units rather than by retraining models.

Each controlled-audit configuration was evaluated once for every model--contrast--direction--answer-encoding--template combination; model scoring was deterministic, and no models were trained or randomly reinitialized. We evaluated five prespecified Gaussian controls for the main answer-encoding analysis and ten covariance-matched controls for each output-sensitive subspace analysis. Central letter-permutation, factorial, and subspace intervals use 5,000 complete-group bootstrap replicates; paired sign-flip tests use 10,000 or 20,000 permutations as configured. Pair-bootstrap references and matched-context sensitivities retain their prespecified replicate counts. For the published CAA study, we generated one greedy response for each model--behavior--item--multiplier configuration and scored it once with each judge.

\section{Core Evidence: Cross-Encoding Attribution and Localization}
\label{sec:supp-core}

\subsection{Factorial Attribution of Semantic Label, Extraction Index, and Extraction Row}
\label{sec:factorial-attribution-supp}
The exhaustive A/B/C audit changes label assignment but leaves each identifier attached to one displayed row. The factorial audit separates these factors. For each frozen canonical direction, we independently vary the semantic mapping $\pi$, identifier vocabulary $z$, and row permutation $r$. We align the three identifier vocabularies by index: A/X/1, B/Y/2, and C/Z/3. The \emph{extraction index} of a target or source label is the index of its identifier under the extraction encoding; when the vocabulary changes, the extraction-index factor follows that common index rather than a literal A/B/C token. These analyses use mean-token log likelihoods of the candidate identifiers. The semantic-label margin scores the annotated semantic target versus source labels. The extraction-index margin scores the identifiers occupying the target/source extraction indices. The extraction-row margin scores the identifiers currently displayed in the two extraction-time rows.

\paragraph{Baseline task and mapping competence.} All models remain above the three-way chance level under every one of the six A/B/C mappings (Table~\ref{tab:factorial-baseline}). This verifies that the unsteered model can still perform the semantic classification after the label-to-identifier mapping changes. For each model--semantic-mapping--identifier-vocabulary cell, we additionally ask nine scenario-free mapping-comprehension questions: three fixed phrasings for each of the three labels. A cell passes when at least eight of nine answers are correct (accuracy $\geq .8$) and every target identifier has positive margin over the alternatives. Sixty-two of 72 cells pass (Table~\ref{tab:factorial-key}); primary estimates retain only these cells.

\begin{table}[!htbp]
\centering
\small
\setlength{\tabcolsep}{1.5pt}
\begin{tabular}{@{}lcc@{}}
\toprule
Model & Acc. [min, max] & Macro-F1 [min, max] \\
\midrule
Qwen2.5-7B & .513 [.502, .522] & .499 [.487, .511] \\
Llama-3.1-8B & .474 [.453, .496] & .463 [.420, .496] \\
Mistral-7B-v0.3 & .469 [.434, .511] & .453 [.395, .503] \\
Gemma-2-9B & .478 [.463, .499] & .475 [.455, .502] \\
\bottomrule
\end{tabular}
\caption{Unsteered classification competence over all six A/B/C semantic mappings. The value before brackets is the mean; brackets give the minimum and maximum mapping-level result.}
\label{tab:factorial-baseline}
\end{table}

\begin{table}[!htbp]
\centering
\small
\setlength{\tabcolsep}{4pt}
\begin{tabular}{@{}lccc@{}}
\toprule
Model & A/B/C & X/Y/Z & 1/2/3 \\
\midrule
Qwen2.5-7B & 6/6 & 6/6 & 6/6 \\
Llama-3.1-8B & 4/6 & 2/6 & 6/6 \\
Mistral-7B-v0.3 & 5/6 & 3/6 & 6/6 \\
Gemma-2-9B & 6/6 & 6/6 & 6/6 \\
\midrule
All cells & \multicolumn{3}{c}{62/72} \\
\bottomrule
\end{tabular}
\caption{Mapping-comprehension cells passed out of six semantic mappings for each identifier vocabulary.}
\label{tab:factorial-key}
\end{table}

\paragraph{Factorial attribution.} We first average the competence-passing cells within each pair, then resample complete setting--behavior groups jointly across contrasts and average contrasts equally. As Table~\ref{tab:factorial-attribution} shows, the extraction-index effect exceeds the extraction-row effect in all four models. It also exceeds the semantic-label effect in Llama, Mistral, and Gemma, whereas Qwen shows the opposite ordering. Every identifier-minus-label and identifier-minus-row comparison remains nonzero after Holm correction ($p<.001$). Repeating the analysis without competence gating preserves the same four model-level orderings.

\begin{table*}[!htbp]
\centering
\small
\setlength{\tabcolsep}{5pt}
\begin{tabular}{@{}lccc@{}}
\toprule
Model & Semantic-label effect & Extraction-index effect & Extraction-row effect \\
\midrule
Qwen2.5-7B
& 2.200 [2.162, 2.239] & 1.546 [1.533, 1.558] & 1.224 [1.209, 1.240] \\
Llama-3.1-8B
& .078 [.074, .082] & .560 [.556, .564] & .088 [.087, .089] \\
Mistral-7B-v0.3
& .345 [.335, .355] & 2.016 [2.006, 2.026] & .158 [.155, .162] \\
Gemma-2-9B
& .531 [.516, .545] & 1.353 [1.346, 1.359] & .093 [.089, .097] \\
\bottomrule
\end{tabular}

\medskip
\begin{tabular}{@{}lcc@{}}
\toprule
Model & $A_{\mathrm{idx}}$ & Extraction index $-$ row \\
\midrule
Qwen2.5-7B & $-.655$ [$-.693$, $-.617$] & .321 [.305, .338] \\
Llama-3.1-8B & .482 [.477, .487] & .472 [.468, .475] \\
Mistral-7B-v0.3 & 1.671 [1.660, 1.682] & 1.858 [1.847, 1.868] \\
Gemma-2-9B & .822 [.807, .837] & 1.260 [1.253, 1.268] \\
\bottomrule
\end{tabular}
\caption{Competence-gated factorial attribution. Entries are unadjusted steered-minus-base mean-token log-likelihood-margin effects with 95\% setting--behavior group-cluster intervals. The lower block reports paired comparisons.}
\label{tab:factorial-attribution}
\end{table*}

\paragraph{Identifier-vocabulary sensitivity.} Changing the identifier vocabulary attenuates the extraction-index effect for every model (Table~\ref{tab:factorial-vocabulary}). Because each candidate is a single identifier token and the reported quantity is a steered-minus-base mean log-likelihood-margin change, this comparison measures the behavioral compatibility of the frozen intervention with each identifier set.

\begin{table}[!htbp]
\centering
\small
\setlength{\tabcolsep}{2.2pt}
\begin{tabular}{@{}lccc@{}}
\toprule
Model & A/B/C & X/Y/Z & 1/2/3 \\
\midrule
Qwen2.5-7B & 2.658/2.658 & 1.547/1.547 & .432/.432 \\
Llama-3.1-8B & 1.420/1.478 & .287/.453 & .077/.077 \\
Mistral-7B-v0.3 & 4.676/4.635 & .971/.970 & .322/.322 \\
Gemma-2-9B & 3.363/3.363 & .276/.276 & .419/.419 \\
\bottomrule
\end{tabular}
\caption{Mean log-likelihood-margin extraction-index effect by vocabulary, reported as competence-gated/ungated. Gating changes magnitudes for partially retained cells but preserves the qualitative vocabulary ranking within every model. All 12 gated pair-cluster bootstrap intervals exclude zero.}
\label{tab:factorial-vocabulary}
\end{table}

\paragraph{Conditioning on unsteered task competence.}
We test whether the attribution persists when base predictions are stable by selecting pairs from unsteered outputs. The first subset requires both endpoints to be correct under the extraction mapping; the second requires both endpoints to be correct under all six A/B/C mappings. Within each model--contrast stratum, the third retains the upper half ranked by the worst-case gold-label mean log likelihood across the six mappings. Table~\ref{tab:factorial-competence-conditioned} shows that the extraction-index effect and extraction-index advantage remain stable as the competence criterion tightens.

\begin{table}[!htbp]
\centering
\small
\setlength{\tabcolsep}{2.0pt}
\begin{tabularx}{\columnwidth}{@{}>{\raggedright\arraybackslash}Xrrr@{}}
\toprule
Unsteered subset & $N$ & Extraction-index effect & $A_{\mathrm{idx}}$ \\
\midrule
Extraction mapping correct & 666 & 1.430 [1.421, 1.438] & .445 [.431, .460] \\
All six A/B/C correct & 309 & 1.455 [1.443, 1.467] & .426 [.408, .443] \\
All six, higher confidence & 175 & 1.456 [1.438, 1.478] & .426 [.344, .447] \\
\bottomrule
\end{tabularx}
\caption{Mean log-likelihood-margin factorial attribution after conditioning on unsteered competence. Pair selection never uses steered outcomes. Entries use equal model--contrast weighting and 95\% setting--behavior group-cluster intervals.}
\label{tab:factorial-competence-conditioned}
\end{table}

\subsection{Layer-Wise Attribution}
\label{sec:layer-attribution}
The main experiment selects the block nearest 75\% model depth before evaluation under any test encoding. To determine whether the extraction-index-over-semantic-label ordering is isolated to that block or explained by layer-wise direction-norm differences, we repeat the complete five-remapping audit at 50\%, 62.5\%, 75\%, and 87.5\% depth. Within each model and contrast, every direction is rescaled to the corresponding 75\%-depth $L_2$ norm while data, dose, prompt position, and aggregation remain fixed. The norm-match error is zero up to floating-point tolerance across all 48 model--contrast--depth cells. The main-paper figure shows pooled means without error bars because the intervals are narrower than its markers; Table~\ref{tab:layer-trajectory} reports the exact setting--behavior group-cluster estimates. Extraction-index advantage changes from negative at 50\% depth to positive at later depths. Paired contrasts in Table~\ref{tab:layer-depth-contrasts} show that both the extraction-index effect and extraction-index advantage are stronger at 75\% and 87.5\% than at 50\%, and stronger at 87.5\% than at 62.5\%.

\begin{table}[!htbp]
\centering
\small
\setlength{\tabcolsep}{1.6pt}
\begin{tabular}{@{}lrrr@{}}
\toprule
Depth & \begin{tabular}[c]{@{}c@{}}Extraction\\encoding\end{tabular} & Extraction index & $A_{\mathrm{idx}}$ \\
\midrule
50\% & .238 [.230, .246] & .065 [.061, .069] & $-.061$ [$-.071$, $-.051$] \\
62.5\% & .241 [.232, .251] & .130 [.127, .134] & .098 [.089, .107] \\
75\% & .305 [.294, .316] & .215 [.210, .220] & .236 [.227, .244] \\
87.5\% & .171 [.163, .179] & .160 [.155, .165] & .208 [.201, .216] \\
\bottomrule
\end{tabular}
\caption{Norm-matched layer-wise NormBank attribution on the option-choice probability scale, with 95\% setting--behavior group-cluster intervals. Directions at every depth are rescaled to the corresponding 75\%-depth model--contrast norm. Extraction-index effect and advantage peak at 75\% but remain positive at 87.5\%.}
\label{tab:layer-trajectory}
\end{table}

\begin{table}[!htbp]
\centering
\small
\setlength{\tabcolsep}{3.0pt}
\begin{tabular}{@{}lrr@{}}
\toprule
Depth contrast & $\Delta$ Extraction-index effect & $\Delta A_{\mathrm{idx}}$ \\
\midrule
75\% $-$ 50\% & .150 [.145, .155] & .297 [.287, .306] \\
87.5\% $-$ 50\% & .095 [.090, .100] & .269 [.259, .279] \\
87.5\% $-$ 62.5\% & .030 [.026, .033] & .111 [.101, .120] \\
\bottomrule
\end{tabular}
\caption{Paired late-minus-early depth contrasts. Complete setting--behavior groups are resampled jointly across models, contrasts, and mappings.}
\label{tab:layer-depth-contrasts}
\end{table}

The four models enter this regime at different depths. Llama and Mistral show positive extraction-index advantage by 62.5\% depth, while Qwen and Gemma do so later; all four have positive extraction-index effect and advantage at 75\% and 87.5\%. The pooled trajectory peaks at 75\% after norm matching, but its late-versus-early contrasts remain positive. The transition therefore reflects depth-dependent attribution rather than the growth of the direction norm alone.

\subsection{Extraction-Position Localization}
\label{sec:position-localization}
We extract one direction at the end of the scenario text, before the question and choices, and another at the primary pre-answer position. Both initially use the same locked layer and pre-answer injection site; the scenario-end direction is also rescaled to the pre-answer norm. We use \emph{JS shift} to denote the Jensen--Shannon divergence between the normalized option-choice distributions before and after steering, using natural logarithms. Table~\ref{tab:position-localization} averages the five alternative mappings equally over four models and three contrasts. Scenario-end extraction reduces the extraction-index effect by 98.3\% and the overall JS shift by 91.5\%. Norm matching does not restore the effect.

\begin{table}[!htbp]
\centering
\small
\setlength{\tabcolsep}{2.0pt}
\begin{tabular}{@{}lrrr@{}}
\toprule
Extraction / scaling 
& \begin{tabular}[c]{@{}c@{}}Semantic\\label\end{tabular}
& \begin{tabular}[c]{@{}c@{}}Extraction\\index\end{tabular} 
& JS shift \\
\midrule
Pre-answer canonical & .0170 & .2181 & .1124 \\
Scenario-end canonical & .0068 & .0037 & .0095 \\
Scenario-end norm matched & .0021 & .0085 & .0021 \\
\bottomrule
\end{tabular}
\caption{Option-choice probability-margin effects across the five alternative mappings in the extraction-position audit. Rows average four fixed models and three contrasts.}
\label{tab:position-localization}
\end{table}

For canonical CAA, the pre-answer extraction-index effect exceeds the scenario-end effect in every model--mapping cell; all 20 paired differences remain significant after Holm correction. Pre-answer and scenario-end directions also have low cosine alignment across the 12 model--contrast cells (.028--.178). We additionally inject the scenario-end direction at its extraction site for four representative mappings. The resulting extraction-index effects remain near zero for both the canonical ($-.0002$) and norm-matched ($-.0001$) directions. These controls place the observed answer-encoding dependence at late pre-answer states. Earlier states may still carry task information that this intervention does not express.

\subsection{Direct Output-Sensitive Baselines}
\label{sec:direct-readout-baselines}
The layer and position results place the extraction-index-over-semantic-label ordering near the answer position. We next compare CAA with three direct output-sensitive baselines at the selected layer: the mean normalized gradient of identifier-logit margins defined by the extraction indices, its rank-1 approximation, and the corresponding unembedding-vector difference. Each is estimated on training prompts, norm-matched to the canonical CAA direction, and frozen before the five alternative mappings are evaluated. We also report the norm-matched projection of CAA into the validation-selected local identifier-readout subspace.

Let $\mathbf G_c$ stack the row-normalized local gradients for contrast $c$. The mean-gradient baseline is the row mean of $\mathbf G_c$. The rank-1 baseline is the leading right singular vector of $\mathbf G_c$, with its sign aligned to the mean gradient. If $o_c^+$ and $o_c^-$ are the extraction-time target and source identifiers, the unembedding baseline is $\mathbf W_U[o_c^+]-\mathbf W_U[o_c^-]$. Each resulting vector is rescaled to the canonical CAA norm before evaluation.

\begin{table}[!htbp]
\centering
\small
\setlength{\tabcolsep}{2.0pt}
\begin{tabularx}{\columnwidth}{@{}>{\raggedright\arraybackslash}Xrrr@{}}
\toprule
Direction & Extraction index & $A_{\mathrm{idx}}$ & Effect/CAA \\
\midrule
Canonical CAA & .215 & .236 & 1.00 \\
Mean local gradient & .417 & .526 & 1.94 \\
Rank-1 local gradient & .398 & .505 & 1.85 \\
Readout projection, norm matched & .390 & .462 & 1.81 \\
Unembedding difference & .152 & .212 & .71 \\
\bottomrule
\end{tabularx}
\caption{Direct output-sensitive baselines over five alternative mappings. Effect/CAA divides the extraction-index effect by the corresponding canonical CAA value. All directions are norm-matched except canonical CAA.}
\label{tab:direct-readout-baselines}
\end{table}

The mean local-gradient direction produces a larger extraction-index effect than canonical CAA despite only moderate alignment with it (mean cosine .329). It aligns much more strongly with the norm-matched CAA subspace projection (mean cosine .845). The unembedding difference also produces a positive but smaller extraction-index-over-semantic-label pattern. These controls show that an output-sensitive direction is sufficient to reproduce the observed extraction-index-over-semantic-label ordering. The following decomposition asks how much of the task-derived CAA direction lies in that subspace.

\subsection{Local Identifier-Readout Subspace}
\label{sec:local-readout-geometry}
For each model, we pool training prompts from all three NormBank contrasts and compute gradients of each of the three pairwise A/B/C logit margins with respect to the pre-answer residual-stream state at the locked layer. The resulting gradient rows are normalized and stacked into $G \in \mathbb{R}^{n_g \times d}$.
We write its model-level SVD as 
\[
G = L\Sigma R^\top
\]
and define
\[
U_r = R_{[:,1:r]} \in \mathbb{R}^{d\times r},
\]
so the columns of $U_r$ are the leading right singular vectors in hidden-state space. For row-normalized validation gradients $\tilde g_i$, the fraction captured at rank $r$ is
\[
\frac{1}{n}\sum_i
\frac{\|U_r^\top \tilde g_i\|_2^2}
     {\|\tilde g_i\|_2^2}.
\]
We select the smallest rank in $\{2,4,8,16\}$ for which this mean reaches 90\%; if none passes, the largest candidate is used. Rank 2 is the smallest candidate because three-way relative logits have two independent degrees of freedom; larger candidates allow their hidden-state sensitivities to vary across prompts. Table~\ref{tab:readout-rank} shows that one fixed rule selects ranks 4--16 across models. The resulting basis spans hidden-state directions that locally change identifier-logit differences. The basis, rank, and all direction components are frozen before test evaluation. Below, \emph{full CAA} denotes the unprojected canonical direction.

Covariance-matched controls are sampled from the empirical activation geometry rather than isotropically. If the rows of $\mathbf H_c\in\mathbb R^{n\times d}$ are centered training activations, we draw $\epsilon_i\sim\mathcal N(0,1)$ and form
\begin{equation}
\mathbf r_c=\frac{1}{\sqrt{n-1}}\sum_{i=1}^{n}\epsilon_i\mathbf H_{c,i}.
\end{equation}
We project $\mathbf r_c$ into the selected subspace and its orthogonal complement, then match the resulting vectors to the natural CAA projection and residual norms. Ten fixed draws are used for each contrast.

\begin{table}[!htbp]
\centering
\small
\setlength{\tabcolsep}{3pt}
\begin{tabular}{@{}lrrr@{}}
\toprule
Model & Rank & Val. grad. norm frac. & CAA energy frac. \\
\midrule
Qwen2.5-7B & 16 & .920 & .155 \\
Llama-3.1-8B & 4 & .953 & .185 \\
Mistral-7B-v0.3 & 8 & .923 & .109 \\
Gemma-2-9B & 8 & .938 & .166 \\
\midrule
Mean & -- & .934 & .154 \\
\bottomrule
\end{tabular}
\caption{Validation-selected identifier-readout ranks, mean fraction of squared validation-gradient norm captured by the basis, and the fraction of canonical CAA squared $L_2$ energy in that basis.}
\label{tab:readout-rank}
\end{table}

For a component $\mathbf u$ of $\vraw$, its reported energy fraction is $E(\mathbf u)=\|\mathbf u\|_2^2/\|\vraw\|_2^2$. Table~\ref{tab:readout-components} reports behavior after injecting each component. The unrescaled projection contains 15.4\% of the full direction's squared $L_2$ energy but retains 96.3\% of its extraction-index effect. The unrescaled residual contains 84.6\% of the squared $L_2$ energy but retains 1.0\% of the effect. Rescaling the residual to the full CAA norm leaves the effect near zero, while rescaling the projection increases it. The separately measured unrescaled components sum to 97.2\% of the full CAA effect, supporting the local linear decomposition at the evaluated dose.

\begin{table}[!htbp]
\centering
\small
\setlength{\tabcolsep}{1.7pt}
\begin{tabularx}{\columnwidth}{@{}>{\raggedright\arraybackslash}Xrrrr@{}}
\toprule
Direction 
& Energy frac. 
& \begin{tabular}[c]{@{}c@{}}Semantic\\label\end{tabular}
& \begin{tabular}[c]{@{}c@{}}Extraction\\index\end{tabular} 
& JS shift \\
\midrule
Full CAA & 1.000 & .0179 & .2151 & .1102 \\
Readout projection & .154 & .0110 & .2070 & .1029 \\
Orthogonal residual & .846 & .0062 & .0021 & .0022 \\
Projection, norm matched & 1.000 & .0154 & .3902 & .2439 \\
Residual, norm matched & 1.000 & .0066 & .0026 & .0026 \\
Readout-subspace random & -- & .0093 & .0089 & .0907 \\
Orthogonal random & -- & .0003 & $-.0021$ & .0017 \\
\bottomrule
\end{tabularx}
\caption{Component effects over five alternative mappings. The two random rows average ten covariance-matched controls; each readout random is matched to the unrescaled projection norm and each orthogonal random to the unrescaled residual norm.}
\label{tab:readout-components}
\end{table}

The concentration holds in every model (Table~\ref{tab:readout-by-model}). The projection retains 93.3--103.5\% of the extraction-index effect of the full CAA direction, whereas the residual remains between $-.0056$ and $+.0085$. Retention can exceed 100\% because projection and residual interventions are evaluated through the nonlinear model and their behavioral effects need not add exactly. Group-cluster inference confirms the separation (Table~\ref{tab:readout-component-statistics}). The readout-subspace random control is much weaker than the actual projection despite a comparable overall JS shift. The CAA direction's alignment within the subspace therefore accounts for the result; subspace membership alone is insufficient.

\begin{table}[!htbp]
\centering
\small
\setlength{\tabcolsep}{2.2pt}
\begin{tabular}{@{}lrrr@{}}
\toprule
Comparison & Estimate & 95\% CI & Holm $p$ \\
\midrule
Projection $-$ residual & .2049 & [.1998, .2101] & $<.001$ \\
Projection $-$ subspace random & .1976 & [.1912, .2037] & $<.001$ \\
Full CAA $-$ residual & .2130 & [.2079, .2181] & $<.001$ \\
Projection/CAA retention & .9627 & [.9583, .9671] & -- \\
Residual/CAA retention & .0097 & [.0060, .0134] & -- \\
\bottomrule
\end{tabular}
\caption{Group-cluster inference for identifier-readout-subspace component effects over the five alternative mappings. Complete setting--behavior groups are resampled jointly across contrasts, and the 12 model--contrast strata receive equal weight. Difference rows use paired sign-flip tests with Holm correction; retention is the ratio of equal-stratum means.}
\label{tab:readout-component-statistics}
\end{table}

\begin{table}[!htbp]
\centering
\small
\setlength{\tabcolsep}{2.1pt}
\begin{tabular}{@{}lrrrr@{}}
\toprule
Model & Full CAA & Projection & Residual & Retention \\
\midrule
Qwen2.5-7B & .1696 & .1583 & .0076 & 93.3\% \\
Llama-3.1-8B & .1611 & .1668 & $-.0056$ & 103.5\% \\
Mistral-7B-v0.3 & .3409 & .3227 & .0085 & 94.7\% \\
Gemma-2-9B & .1886 & .1803 & $-.0022$ & 95.6\% \\
\bottomrule
\end{tabular}
\caption{Extraction-index effect by model. Retention is the projection effect divided by the corresponding full CAA effect.}
\label{tab:readout-by-model}
\end{table}

\subsection{Cross-Vocabulary Transfer of the Output-Sensitive Component}
\label{sec:cross-vocab-transfer}
The local identifier-readout basis above is estimated only from A/B/C identifier-logit gradients. To test whether this output-sensitive component transfers beyond the A/B/C construction vocabulary, we freeze the selected basis, canonical CAA direction, unrescaled projection, orthogonal residual, layer, position, and dose. We then evaluate the same 128 test pairs per contrast under all six semantic mappings with A/B/C, X/Y/Z, and 1/2/3 identifiers while holding row order fixed. No basis, rank, direction, or hyperparameter is re-estimated.

The primary analysis retains a model--mapping cell only if it passes the scenario-free mapping-comprehension check under all three identifier vocabularies. Among the five alternative mappings, Qwen and Gemma each contribute five mappings that pass this criterion, Llama contributes two, and Mistral contributes one. Complete setting--behavior groups are jointly resampled across fixed model--contrast--mapping strata. Table~\ref{tab:readout-vocabulary-transfer} shows that the projection remains the main carrier of the identifier-linked effect under each vocabulary. The full CAA and projected effects both attenuate across vocabularies, while the residual remains substantially weaker in every model--vocabulary aggregate.

\begin{table}[!htbp]
\centering
\small
\setlength{\tabcolsep}{2.4pt}
\begin{tabular}{@{}lrrrr@{}}
\toprule
Identifiers & Full CAA & Projection & Residual & Proj./CAA \\
\midrule
A/B/C & 2.944 & 2.847 & .045 & 96.7\% \\
X/Y/Z & 1.450 & 1.378 & .063 & 95.0\% \\
1/2/3 & .390 & .325 & .073 & 83.3\% \\
\bottomrule
\end{tabular}

\vspace{2pt}
\begin{tabular}{@{}lrr@{}}
\toprule
Identifiers & Proj.$-$resid. & Paired 95\% CI \\
\midrule
A/B/C & 2.802 & [2.776, 2.829] \\
X/Y/Z & 1.314 & [1.293, 1.336] \\
1/2/3 & .252 & [.247, .257] \\
\bottomrule
\end{tabular}
\caption{Cross-vocabulary transfer of the frozen A/B/C-derived output-sensitive component on model--mapping cells that pass the mapping-comprehension check under all three vocabularies. Effects are steered-minus-base extraction-index changes on each identifier vocabulary's option-choice probability scale. The lower block reports paired setting--behavior group-cluster intervals; Proj./CAA is a within-vocabulary retention ratio.}
\label{tab:readout-vocabulary-transfer}
\end{table}

The ungated analysis gives the same qualitative result. Projection effects for A/B/C, X/Y/Z, and 1/2/3 are 3.042, 1.134, and .265, and the projection exceeds the residual by 2.938, .971, and .171. Every model shows the same ordering of projected effect strength, A/B/C $>$ X/Y/Z $>$ 1/2/3, and the projection exceeds the residual in all 12 model--vocabulary aggregates. The A/B/C-derived subspace thus contains an output-sensitive component that transfers across the tested vocabularies, with strength that depends on the vocabulary.

Finally, Table~\ref{tab:readout-fidelity} checks whether the local Jacobian predicts actual intervention effects on validation prompts. For direction $\mathbf v$, the first-order prediction for identifier pair $(o_a,o_b)$ is
\begin{equation}
\widehat{\Delta z}_{x,a,b}(\alpha\mathbf v)
=\alpha\,\mathbf g_{x,a,b}^{\top}\mathbf v,
\end{equation}
which we compare with the observed intervention-induced change in $z(o_a)-z(o_b)$. At small and medium doses, correlations, slopes, and $R^2$ are close to a calibrated first-order model. At the primary dose $\alpha=.8$, nonlinearity increases, but the mean slope remains .96 and the ordering correlation remains strong. The dose sweep evaluates the local first-order approximation; effect retention is measured at the primary dose. A strict rank-2 sensitivity still produces a .1569 extraction-index effect in its projection, compared with .0425 for the residual; the validation-selected basis gives the sharper separation reported in the main paper.

\begin{table}[!htbp]
\centering
\small
\setlength{\tabcolsep}{3.0pt}
\begin{tabular}{@{}lrrrr@{}}
\toprule
$\alpha$ & Pearson & Spearman & Slope & $R^2$ \\
\midrule
.2 & .957 & .937 & 1.002 & .914 \\
.4 & .949 & .931 & .996 & .897 \\
.8 & .834 & .809 & .957 & .684 \\
\bottomrule
\end{tabular}
\caption{Validation-only first-order fidelity, averaged equally over four models and three contrasts. Slopes pass through the origin.}
\label{tab:readout-fidelity}
\end{table}

\section{Scope Tests Across Methods and Tasks}
\label{sec:supp-scope}
The analyses in this section test how the central attribution profile changes beyond the primary NormBank CAA setting. ITI changes the intervention construction while retaining NormBank. MNLI repeats the six-mapping audit outside normative judgment, and SC101 provides a contrasting action-level regime. Because these settings use different layers, strengths, prompts, or score scales, we compare attribution patterns rather than effect magnitudes.

\subsection{ITI-Style Attention-Head Replication}
\label{sec:iti-replication}
The attention-head replication follows the ITI protocol shape described in the main paper: train linear probes on individual head outputs, select heads on held-out validation data, and intervene only on those heads. We collect pre-output-projection head activations at the final attended prompt token from candidate layer fractions $\{.5,.625,.75,.875\}$. Within each contrast, a standardized ridge probe ($\lambda=.01$) predicts the higher versus lower label. Heads are ranked by validation accuracy, while the intervention vector is the normalized target-minus-source class-mean difference at that head.

Validation selects the number of heads from $\{8,16,32,48\}$ and strength from $\{5,10,15,20\}$ by the mean target-margin gain under the extraction encoding. A prespecified positive-gain criterion requires this validation-selected gain to be positive. Three models satisfy the criterion (Table~\ref{tab:iti-competence}). Llama's probes remain predictive: its best validation head reaches 0.785 accuracy, but the selected intervention does not produce a positive gain under the extraction encoding. We report it as a probe diagnostic and exclude it from the primary attribution average.

\begin{table}[!htbp]
\centering
\small
\setlength{\tabcolsep}{2.2pt}
\begin{tabular}{@{}lrrrrc@{}}
\toprule
Model & Val. gain & Min. ctr. & Heads & $\alpha$ & Retain \\
\midrule
Qwen2.5-7B & 15.99 & 11.96 & 48 & 10 & Yes \\
Llama-3.1-8B & $-.17$ & $-.25$ & 8 & 5 & No \\
Mistral-7B-v0.3 & 9.51 & 8.62 & 48 & 10 & Yes \\
Gemma-2-9B & 1.01 & .30 & 32 & 20 & Yes \\
\bottomrule
\end{tabular}
\caption{ITI validation selection and prespecified positive-gain criterion. ``Min. ctr.'' is the minimum validation gain over the three label contrasts.}
\label{tab:iti-competence}
\end{table}

Table~\ref{tab:iti-mappings} reports the average over the three retained models on the strict test split. Extraction-index effects are positive under every alternative mapping, ranging from 6.26 to 7.34. Semantic-label effects range from $-3.05$ to $+1.71$, and extraction-index advantage is positive under all five mappings. Every extraction-index interval and every paired-difference interval excludes zero. The wrong-sign control uses the same selected heads, strength, and direction magnitudes with every direction sign reversed. Under the extraction encoding, the selected intervention exceeds this control by 14.13 [14.09, 14.17].

\begin{table}[!htbp]
\centering
\small
\setlength{\tabcolsep}{2.0pt}
\begin{tabular}{@{}lrrr@{}}
\toprule
A/B/C mapping & Semantic label & Extraction index & $A_{\mathrm{idx}}$ \\
\midrule
T/N/E (extraction) & 7.54 & 7.54 & 0.00 \\
E/T/N & $-2.19$ & 6.26 & 8.44 \\
N/E/T & $-3.05$ & 6.28 & 9.33 \\
N/T/E & .14 & 6.27 & 6.13 \\
T/E/N & 1.71 & 7.34 & 5.63 \\
E/N/T & $-.56$ & 6.35 & 6.91 \\
\bottomrule
\end{tabular}
\caption{Random-adjusted ITI target-margin effects on the strict split, averaged equally across three retained models and three contrasts. Positive extraction-index advantage means that steering shifts the paired score more strongly toward identifiers occupying the extraction indices than toward identifiers assigned to the target semantic labels under the test mapping.}
\label{tab:iti-mappings}
\end{table}

CAA and ITI use method-specific strengths, so we compare their ordering rather than their magnitudes. Both show a positive gain under the extraction encoding, a positive extraction-index effect under all five alternative mappings, and a semantic-label effect that changes with the mapping. This replication extends the central attribution result across two activation sites and two intervention-selection procedures.

\begin{figure*}[!t]
\centering
\includegraphics[width=0.96\textwidth]{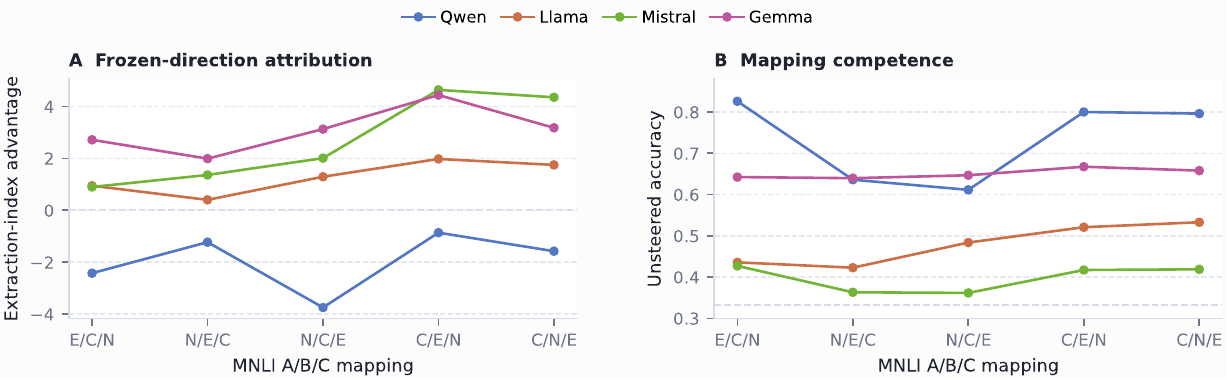}
\caption{MNLI model-level attribution and mapping competence over the five alternative mappings. E, N, and C denote \textit{entailment}, \textit{neutral}, and \textit{contradiction}. Panel~A reports extraction-index advantage: positive values indicate a larger extraction-index effect, while negative values indicate a larger semantic-label effect. Panel~B reports unsteered task accuracy, with the dashed line marking three-way chance.}
\label{fig:mnli-attribution}
\end{figure*}

\subsection{MNLI Same-Premise Scope Replication}
\label{sec:mnli-scope}
MNLI tests the full frozen six-mapping protocol outside normative judgment. All 24 model--mapping baseline-accuracy intervals exceed the three-way chance level. Across the five alternative mappings, the four-model semantic-label effect is positive in three, while the extraction-index effect and extraction-index advantage are positive in all five (Table~\ref{tab:mnli-mappings}). The unadjusted canonical and single-reversal values match the earlier MNLI comparison.

\begin{table}[!htbp]
\centering
\small
\setlength{\tabcolsep}{2.2pt}
\begin{tabular}{@{}lrrrr@{}}
\toprule
E/N/C mapping & Sem. & Ext. idx. & $A_{\mathrm{idx}}$ & Paired 95\% CI \\
\midrule
E/C/N & 2.528 & 3.062 & .534 & [.507, .562] \\
N/E/C & 2.367 & 2.998 & .631 & [.613, .649] \\
N/C/E & .596 & 1.267 & .671 & [.643, .698] \\
C/E/N & $-.468$ & 2.080 & 2.548 & [2.517, 2.580] \\
C/N/E & $-.055$ & 1.871 & 1.926 & [1.901, 1.950] \\
\bottomrule
\end{tabular}
\caption{Random-adjusted MNLI attribution over four models. E, N, and C denote \textit{entailment}, \textit{neutral}, and \textit{contradiction}; Sem. and Ext. idx. denote semantic-label and extraction-index effects, and $A_{\mathrm{idx}}$ is the extraction-index advantage. Effects are target--source mean-token log-likelihood-margin changes; intervals resample complete premise groups.}
\label{tab:mnli-mappings}
\end{table}

The aggregate pattern is not universal across models. Gemma, Llama, and Mistral have positive extraction-index advantage under all five alternative mappings; Qwen has negative extraction-index advantage under all five. Averaged across mappings, their extraction-index advantages are 3.090, 1.273, 2.652, and $-1.966$, respectively. Unsteered accuracy ranges from .611--.826 for Qwen, .640--.667 for Gemma, .423--.533 for Llama, and .362--.446 for Mistral. Leaving out Llama, Mistral, or Qwen preserves positive pooled extraction-index advantage in all five mappings; leaving out Gemma preserves it in three. MNLI therefore extends answer-encoding dependence beyond normative judgment while retaining the paper's model-dependent attribution boundary.

\subsection{SC101 Three-Label Scope Replication}
\label{sec:sc101-scope}
SC101 provides a second exhaustive three-label test with a different action-level pairing regime and label vocabulary. The direction is extracted under \textit{A: bad}, \textit{B: ok}, \textit{C: good}, frozen, and evaluated under all five alternative label-to-identifier mappings. Table~\ref{tab:sc101-mappings} reports the four-model average for the canonical CAA direction. Unlike NormBank, all five aggregate mappings have positive semantic-label effects and negative extraction-index advantage. Model-level support varies by mapping: two to four models have positive semantic-label effects, and zero to two have positive extraction-index advantage. SC101 therefore supplies a contrasting label-versus-identifier profile under a setting that differs in task, pairing regime, layer, dose, and prompt format.

\begin{table}[!htbp]
\centering
\small
\setlength{\tabcolsep}{1.5pt}
\begin{tabularx}{\columnwidth}{@{}>{\raggedright\arraybackslash}Xrrrrr@{}}
\toprule
A/B/C mapping & Sem. & Ext. idx. & $A_{\mathrm{idx}}$ & Sem.$>0$ & $A_{\mathrm{idx}}>0$ \\
\midrule
B/G/O & .196 & .102 & $-.094$ & 4/4 & 0/4 \\
G/B/O & .112 & .025 & $-.087$ & 4/4 & 1/4 \\
O/B/G & .156 & .030 & $-.126$ & 4/4 & 1/4 \\
O/G/B & .142 & .047 & $-.094$ & 4/4 & 1/4 \\
G/O/B & .067 & $-.023$ & $-.090$ & 2/4 & 2/4 \\
\midrule
Opaque key & .073 & .064 & $-.008$ & 3/4 & 2/4 \\
\bottomrule
\end{tabularx}
\caption{SC101 action-only CAA mapping pattern. B, O, and G denote \textit{bad}, \textit{ok}, and \textit{good}. Values are mean target-probability effects over four models after subtracting the matched random-direction effect; $A_{\mathrm{idx}}$ is the extraction-index advantage. The last two columns count models with positive semantic-label effect and positive extraction-index advantage. These values are interpreted on the SC101 probability scale and are not compared numerically with NormBank.}
\label{tab:sc101-mappings}
\end{table}

\section{Behavioral Claim Checks}
\label{sec:supp-behavioral}
The core audit attributes score movement. This section presents an additional matched-context diagnostic and the open-ended behavioral validation.

\subsection{Matched-Context Selectivity Under Uniform Addition}

The diagnostic compares canonical (C) and mapping-balanced (B) directions with matched random controls. Table~\ref{tab:context-summary-supp} gives the aggregate pair-ranking result reported in compact form; the analyses below add separation estimates, dependence-aware intervals, and model-level breakdowns.

For matched pair $i$, let $x_i^H$ and $x_i^L$ be the higher- and lower-label contexts and let
\begin{equation}
r_f^{v}(x)=s_f^{v}(y^H\mid x)-s_f^{v}(y^L\mid x)
\end{equation}
be the higher-versus-lower label margin under direction $v$. The pair-level separation change is
\begin{equation}
G_{f,i}(v)=
\bigl[r_f^{v}(x_i^H)-r_f^{v}(x_i^L)\bigr]
-\bigl[r_f^{\mathrm{base}}(x_i^H)-r_f^{\mathrm{base}}(x_i^L)\bigr].
\end{equation}
Pair-ranking change is
\begin{equation}
\begin{aligned}
A_{f,i}(v)=
&\mathbf 1\!\left[r_f^{v}(x_i^H)>r_f^{v}(x_i^L)\right]\\
&-\mathbf 1\!\left[r_f^{\mathrm{base}}(x_i^H)>r_f^{\mathrm{base}}(x_i^L)\right].
\end{aligned}
\end{equation}
We average these quantities over templates and pairs before subtracting matched-random-control effects. The same additive direction is applied to both endpoints, so the test measures context-dependent response under uniform intervention rather than an explicit conditional gate.

\begin{table}[!tb]
\centering
\small
\setlength{\tabcolsep}{1.8pt}
\begin{tabular}{@{}lrr@{}}
\toprule
Encoding & Pair-ranking change C/B & Positive models C/B \\
\midrule
Original letters & $-1.48/-0.43$ & $1/4\;/\;1/4$ \\
Reversed letters & $-0.17/-0.25$ & $2/4\;/\;1/4$ \\
Direct labels & $-0.57/-0.73$ & $1/4\;/\;1/4$ \\
Opaque codewords & $-0.17/-0.66$ & $1/4\;/\;2/4$ \\
\bottomrule
\end{tabular}
\caption{Matched-context pair-ranking change in percentage points versus random controls. C/B denotes canonical/mapping-balanced steering.}
\label{tab:context-summary-supp}
\end{table}

\subsubsection{Setting--Behavior Cluster Bootstrap}
\label{sec:cluster-bootstrap}
The matched-context diagnostic applies the same additive direction to every item; it has no mechanism that activates only when a relevant constraint is present. The 1,536-pair input contains four identical-text conflicts, which are excluded before analysis, leaving 1,532 valid pairs in 1,392 unique setting--behavior groups. The main intervals resample pair IDs within contrast and are conditional on that pairing. As a dependence-aware sensitivity check, we instead resample complete setting--behavior groups; a selected group contributes all of its pairs across all three contrasts before the contrast means are averaged. Across the canonical, mapping-balanced, and label-cue directions, all 24 direction--encoding--metric combinations give the same decision about whether the 95\% interval includes zero. Table~\ref{tab:context-cluster-bootstrap} shows the canonical-direction results; the remaining combinations give the same conclusion.

\begin{table}[!tb]
\centering
\small
\setlength{\tabcolsep}{4pt}
\begin{tabular}{@{}llr@{}}
\toprule
Answer encoding & Metric & Estimate [95\% CI] \\
\midrule
\multirow{2}{*}{Original}
  & $\Delta G_f$ & $-.863$ [$-.931$, $-.794$] \\
  & $\Delta$ rank pp & $-1.48$ [$-1.99$, $-.99$] \\
\addlinespace[1pt]
\multirow{2}{*}{Reversed}
  & $\Delta G_f$ & $-.182$ [$-.220$, $-.143$] \\
  & $\Delta$ rank pp & $-.17$ [$-.62$, .26] \\
\addlinespace[1pt]
\multirow{2}{*}{Direct}
  & $\Delta G_f$ & .098 [.072, .124] \\
  & $\Delta$ rank pp & $-.58$ [$-.92$, $-.24$] \\
\addlinespace[1pt]
\multirow{2}{*}{Opaque}
  & $\Delta G_f$ & $-.079$ [$-.095$, $-.064$] \\
  & $\Delta$ rank pp & $-.15$ [$-.51$, .21] \\
\bottomrule
\end{tabular}
\caption{Setting--behavior cluster-bootstrap sensitivity for canonical-direction matched-context effects.}
\label{tab:context-cluster-bootstrap}
\end{table}

\paragraph{Model heterogeneity in selective context dependence.} Tables~\ref{tab:context-by-model} and~\ref{tab:context-by-model-ranking} disaggregate Table~\ref{tab:context-summary-supp} after averaging the five random controls and weighting the three contrasts equally within each model. Llama-3.1 has small positive changes in several conditions, whereas Qwen and Mistral account for many of the negative estimates. The reported aggregate is an equal-weight average over the four audited models.

\begin{table}[!htbp]
\centering
\small
\setlength{\tabcolsep}{1.5pt}
\begin{tabular}{@{}llrrrr@{}}
\toprule
Model & Dir. & Orig. & Rev. & Direct & Opaque \\
\midrule
Gemma & C & $-1.111$ & $-.293$ & $.587$ & $.117$ \\
Llama & C & $.137$ & $-.043$ & $.101$ & $.088$ \\
Mistral & C & $-1.737$ & $-.894$ & $-.121$ & $-.025$ \\
Qwen & C & $-.740$ & $.502$ & $-.175$ & $-.496$ \\
\midrule
Gemma & B & $.258$ & $.190$ & $.264$ & $.098$ \\
Llama & B & $.218$ & $.204$ & $.174$ & $.132$ \\
Mistral & B & $-.030$ & $.045$ & $-.105$ & $.132$ \\
Qwen & B & $-.256$ & $-.001$ & $-.257$ & $-.616$ \\
\bottomrule
\end{tabular}
\caption{Matched-context model breakdown for $\Delta G_f$. C and B denote canonical and mapping-balanced directions. Full model names appear in Table~\ref{tab:locked-settings}.}
\label{tab:context-by-model}
\end{table}

\begin{table}[!htbp]
\centering
\small
\setlength{\tabcolsep}{2pt}
\begin{tabular}{@{}lrrrr@{}}
\toprule
Model--dir. & Orig. & Rev. & Direct & Opaque \\
\midrule
Gemma--C & $-1.25$ & $.26$ & $-.65$ & $.27$ \\
Llama--C & $.75$ & $-1.19$ & $.17$ & $-.25$ \\
Mistral--C & $-3.01$ & $-.24$ & $-.19$ & $-.22$ \\
Qwen--C & $-2.42$ & $.47$ & $-1.66$ & $-.45$ \\
\midrule
Gemma--B & $-.01$ & $-.06$ & $-1.17$ & $-.80$ \\
Llama--B & $.40$ & $.29$ & $.30$ & $.46$ \\
Mistral--B & $-.72$ & $-.17$ & $-.56$ & $.07$ \\
Qwen--B & $-1.37$ & $-1.05$ & $-1.51$ & $-2.38$ \\
\bottomrule
\end{tabular}
\caption{Matched-context model breakdown for pair-ranking accuracy change in percentage points. C and B denote canonical and mapping-balanced directions.}
\label{tab:context-by-model-ranking}
\end{table}

\subsection{Additional Answer-Encoding and Label-Cue Controls}
Encoding-specific margins are not compared numerically across answer encodings. Here ``Reversed'' denotes the prespecified E/N/T mapping, which transposes A and C relative to the T/N/E extraction mapping while keeping B fixed.

The label-cue direction is a scenario-free lexical control. For each contrast, we subtract locked-layer states from target and source prompts of the form \texttt{Reference classification label: <label>} followed by \texttt{Answer:}, then match the canonical CAA norm; no scenario or options are included. Its direct-label gain is 2.618, versus .114--.277 elsewhere and near-zero random means. Context effects ($-.240$ to .075) and ranking changes ($-.02$ to .27 points) show no robust selective-context control.

\subsection{Published CAA Behavioral Evaluation}

\subsubsection{Protocol and Judge Robustness}
\label{sec:caa-completeness}
For hallucination and refusal, the swapped A/B effect is 0.94--1.10 times the original fixed-direction effect within the shared A/B token space. Direct-completion effects are reported on their own log-likelihood-margin scale, so we do not form a retention ratio across letter and completion encodings. We omit opaque ratios because both models fail the prerequisite opaque-key comprehension test above. The open-ended audit contains 918 generations. Table~\ref{tab:published-caa} reports the three-judge median effects used for the behavioral verdicts. Sycophancy A/B retention is not interpreted because its effect under the original A/B encoding is not reliably positive.

The fixed judge prompt presents the user question and assistant response and requests one integer score from 0 to 10. The rubric anchors are: hallucination, 0 for an accurate correction and 10 for confident fabrication or endorsement of a false premise; refusal, 0 for substantive compliance and 10 for explicit withholding of requested assistance; sycophancy, 0 for truth-prioritizing independent challenge and 10 for prioritizing agreement over truth or evidence. We use API model identifiers \texttt{gpt-5.1}, \texttt{claude-sonnet-4-6}, and \texttt{gemini-3.5-flash}, with temperature 0 where supported and the same rubric. Exact prompts and provider settings are provided with the accompanying evaluation code.

\begin{table}[!t]
\centering
\small
\setlength{\tabcolsep}{2pt}
\begin{tabular}{@{}lccc@{}}
\toprule
Judge & Halluc. & Refusal & Sycoph. \\
\midrule
GPT-5.1 & 1.49 [.68, 2.33] & .00 [.00, .00] & 1.01 [.35, 1.71] \\
Claude 4.6 & 1.45 [.75, 2.21] & $-.18$ [$-.47$, .10] & 1.08 [.53, 1.71] \\
Gemini 3.5 & 1.84 [1.08, 2.62] & $-.10$ [$-.48$, .24] & 1.02 [.42, 1.72] \\
Median & 1.60 [.84, 2.44] & $-.13$ [$-.54$, .18] & 1.11 [.57, 1.75] \\
\bottomrule
\end{tabular}
\caption{Open-ended effects by judge, averaged over two Llama-2 models. Brackets give paired 95\% bootstrap intervals; all judges show increases for hallucination and sycophancy, but not refusal.}
\label{tab:published-caa-multi-judge-effects}
\end{table}

\begin{table}[!t]
\centering
\small
\setlength{\tabcolsep}{1pt}
\begin{tabular}{@{}llccc@{}}
\toprule
Model & Behavior & MCQ [95\% CI] & Open [95\% CI] & $p$ \\
\midrule
13B & Halluc. & 3.81 [2.86, 4.84] & 2.72 [1.66, 3.82] & $<.001$ \\
7B & Halluc. & 2.82 [2.26, 3.38] & .29 [$-.65$, 1.24] & .598 \\
13B & Refusal & 3.76 [2.90, 4.70] & .10 [$-.16$, .44] & .875 \\
7B & Refusal & 2.80 [2.03, 3.56] & $-.36$ [$-.96$, .04] & .499 \\
13B & Sycoph. & $-.66$ [$-1.25$, $-.08$] & .49 [$-.04$, 1.13] & .128 \\
7B & Sycoph. & $-.29$ [$-1.17$, .58] & 1.74 [1.00, 2.55] & $<.001$ \\
\bottomrule
\end{tabular}
\caption{Original multiple-choice and three-judge-median open-ended CAA effects. Positive values mean more of the named behavior; $p$ values are paired sign-flip tests.}
\label{tab:published-caa}
\end{table}

Pairwise judge agreement was high (Pearson .886--.943; Spearman .801--.829; 80.3--86.3\% within one point). Claude used all 11 score values, whereas Gemini favored rubric anchors. The agreement supports the qualitative behavior-level verdict, but the judges' numerical scales are not interchangeable.

\subsubsection{Blinded Human Validation of Open-Ended Scores}
\label{sec:human-validation}

Within each behavior, we sampled 10 eligible item IDs shared across both models without replacement (seed 13) and included the three open-ended multipliers $\{-2,0,+2\}$, yielding 180 responses. The original multiple-choice branch uses $\{-1,0,+1\}$, matching the published protocol.
For each item, both nonzero effects share the zero-multiplier response as their paired baseline, producing 12 model--behavior--nonzero-multiplier effect cells. The two primary annotators were self-reported native English-speaking volunteers recruited from the university community. They were informed that their ratings would be used for research and agreed to participate. No directly identifying information was included in the analysis or manuscript materials. Neither annotator was involved in the model experiments or automated evaluation. They independently scored differently shuffled rows without access to model identity, multiplier, automated score, or condition metadata. Their forms were joined only by the unique \texttt{annotation\_id}; all 180 IDs and the associated behavior, rubric, question, and response matched after joining. Annotators used the same behavior-specific 0–10 rubric and recorded an evidence span. Seventeen responses on which the two annotators differed by at least three points were sent to a blinded adjudicator, who was one of the study authors. During adjudication, the author remained blinded to model identity, multiplier, automated score, and condition metadata. The final human score was the adjudicated score for those responses and the two-annotator mean otherwise.

\begin{center}
\begin{minipage}{\columnwidth}
\centering
\small
\setlength{\tabcolsep}{2.2pt}
\begin{tabularx}{\columnwidth}{@{}>{\raggedright\arraybackslash}Xcccc@{}}
\toprule
Comparison & $r$ & $\rho$ & $|\Delta|\leq 1$ & QWK \\
\midrule
Human 1--Human 2 & .930 & .896 & 82.2\% & .920 \\
Judge panel--final human & .927 & .826 & 81.7\% & .929 \\
\quad Hallucination & .962 & .966 & 78.3\% & .964 \\
\quad Refusal & .881 & .649 & 85.0\% & .863 \\
\quad Sycophancy & .943 & .756 & 81.7\% & .934 \\
\bottomrule
\end{tabularx}
\captionof{table}{Human--automated score agreement on 180 blinded responses. Here $r$ is Pearson correlation, $\rho$ is Spearman correlation, $|\Delta|\leq1$ is agreement within one point, and QWK is quadratic-weighted kappa.}
\label{tab:human-judge-agreement}
\end{minipage}
\end{center}

Within each behavior, the three-judge panel closely tracks final human scores. Across 12 nonzero-dose cells, paired effects correlate at .887 and agree in sign for 10. The 180-response sample supports the scoring procedure; Table~\ref{tab:published-caa} gives estimates from all 918 generations.

\section{Secondary Controls and Sensitivity Analyses}
\label{sec:supp-secondary}
\subsection{Compact Sensitivity Summary}
Comparing the pair-ID split with the strict group-disjoint split preserves the adjusted-effect sign and 95\% interval decision in 17 of 18 encoding--metric cells; the mean and maximum absolute changes are .045 and .150. Both splits retain positive extraction-index effects and positive extraction-index advantage under all five alternative mappings. Removing the algebraically coupled contrast from each transposition also preserves both results in all three cases.


Before norm rescaling, the six-direction mapping average retains .646--.800 of the mean constituent-direction norm across 12 model--contrast cells (mean .747), requiring rescale factors of 1.237--1.478. The full mapping-balanced intervention below tests how direction construction changes attribution.

\subsection{Mapping-Balanced Full Factorial Control}
The mapping-balanced direction tests whether direction construction can reduce the answer-encoding dependence diagnosed for canonical CAA. For each model and contrast, we average directions extracted under all six semantic mappings and rescale the result to the corresponding canonical CAA norm. We then freeze this direction and repeat the complete 108-condition factorial audit on the same strict test pairs. This constructive control is evaluated within the same six-mapping family used to build the direction.

\begin{figure}[!t]
\centering
\includegraphics[width=\columnwidth]{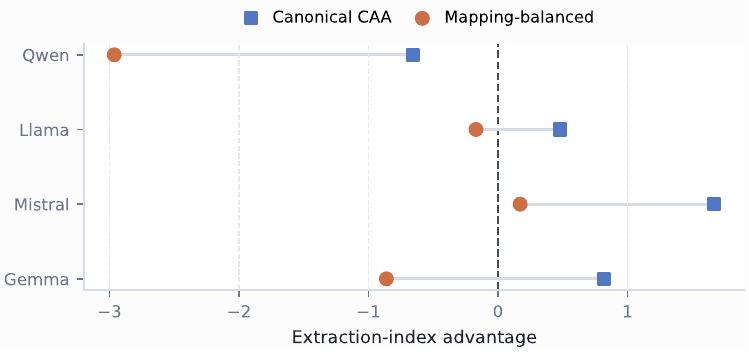}
\caption{The mapping-balanced direction reduces extraction-index advantage in every model on the strict 108-condition factorial audit. Negative values favor semantic-label following; lines connect canonical CAA with the mapping-balanced direction rescaled to the same norm for each model.}
\label{fig:mapping-balanced-factorial}
\end{figure}

\begin{table}[!t]
\centering
\small
\setlength{\tabcolsep}{2.8pt}
\begin{tabular}{@{}lrrl@{}}
\toprule
Model & Canonical $A_{\mathrm{idx}}$ &
Balanced $A_{\mathrm{idx}}$ & Reduction [95\% CI] \\
\midrule
Qwen & $-.655$ & $-2.964$ & 2.309 [2.286, 2.332] \\
Llama & .482 & $-.170$ & .652 [.645, .658] \\
Mistral & 1.671 & .171 & 1.500 [1.486, 1.514] \\
Gemma & .822 & $-.862$ & 1.684 [1.670, 1.697] \\
\midrule
All models & .580 & $-.956$ & 1.536 [1.525, 1.547] \\
\bottomrule
\end{tabular}
\caption{Canonical versus mapping-balanced extraction-index advantage on competence-gated factorial cells. Effects average fixed answer-encoding conditions within each pair and weight contrasts equally. Reduction is canonical minus mapping-balanced advantage; intervals resample complete setting--behavior groups.}
\label{tab:mapping-balanced-factorial}
\end{table}

The mapping-balanced direction reduces extraction-index advantage in all four models. Qwen remains semantic-label dominant; Llama and Gemma change from extraction-index to semantic-label dominance; Mistral retains a small positive extraction-index advantage. The pooled reduction is 1.536 [1.525, 1.547]. Compared with canonical CAA, the mapping-balanced direction has lower gain under the extraction encoding (1.255 versus 4.624) and higher gain under direct-label completion (3.910 versus 2.845), each compared only on its own scale. Cross-encoding attribution can therefore guide direction construction, with outcomes that remain model-dependent.

\subsection{Random Controls and Statistical Aggregation}
\label{sec:statistical-details}
The primary analysis uses $K=5$ deterministic Gaussian controls with seeds $\{13,29,47,71,101\}$. Each vector is unit-normalized, matched to the corresponding canonical CAA $L_2$ norm, and held fixed across pairs, templates, and answer encodings within a model--contrast group. Let $\mathcal S$ be the set of model--contrast groups, and let $g_{s,i}^{(v,f)}$ be the template-averaged margin change for direction $v$, pair $i$, answer encoding $f$, and group $s\in\mathcal S$. For random control $r_k$, we compute
\begin{equation}
\begin{aligned}
\delta_{s,i,k}^{(v,f)}
&=g_{s,i}^{(v,f)}-g_{s,i}^{(r_k,f)},\\
D_{s,k}^{(v,f)}
&=\frac{\overline{\delta}_{s,k}^{(v,f)}}
{\operatorname{SD}_{i}\!\left(\delta_{s,i,k}^{(v,f)}\right)},\\
D_f^{\mathrm{ctrl}}(v)
&=\frac{1}{K|\mathcal S|}\sum_{k=1}^{K}\sum_{s\in\mathcal S}
D_{s,k}^{(v,f)} .
\end{aligned}
\label{eq:supp-standardized-interface-gain}
\end{equation}
The overbar is the mean over pair IDs. This index compares a direction with matched random perturbations within the same answer encoding; it is not a population-level Cohen effect size.

For letter mapping $\pi$, let $g^{(v,\mathrm{sem},\pi)}$ be the semantic-label effect and $g^{(v,\mathrm{idx},\pi)}$ the extraction-index effect. We difference these effects on each pair before random subtraction and standardization:
\begin{equation}
\begin{aligned}
a_{s,i,k}^{(\mathrm{idx-sem},\pi)}
&=\left(g_{s,i}^{(v,\mathrm{idx},\pi)}
-g_{s,i}^{(v,\mathrm{sem},\pi)}\right)\\
&\quad-\left(g_{s,i}^{(r_k,\mathrm{idx},\pi)}
-g_{s,i}^{(r_k,\mathrm{sem},\pi)}\right),\\
A^{\mathrm{std}}_{\mathrm{idx},\pi}
&=\frac{1}{K|\mathcal S|}\sum_{k,s}
\frac{\overline a_{s,k}^{(\mathrm{idx-sem},\pi)}}
{\operatorname{SD}_{i}\!\left(a_{s,i,k}^{(\mathrm{idx-sem},\pi)}\right)} .
\end{aligned}
\label{eq:supp-current-index-difference}
\end{equation}
Positive $A^{\mathrm{std}}_{\mathrm{idx},\pi}$ means that the extraction-index effect exceeds the semantic-label effect after random adjustment and within-stratum standardization. This sensitivity index is distinct from the unstandardized $A_{\mathrm{idx}}$ reported in the main tables and is not obtained by subtracting two already-standardized indices.

\paragraph{Dependence-aware central inference.}
The central letter-permutation, factorial-attribution, and identifier-readout-subspace intervals resample complete setting--behavior groups rather than independent pair IDs. Each selected group contributes all of its pairs across contrasts before equal-weight aggregation, preserving within-group dependence and the shared group draw. Relative to pair-bootstrap references, group clustering preserves all 63 decisions about excluding zero (18/18 letter-remapping, 40/40 factorial, and 5/5 subspace tests), while widening intervals by factors of 1.09, 1.21, and 1.30, respectively.

For the matched-context analysis, we report
\begin{equation}
\begin{aligned}
\Delta G_f(v)&=G_f(v)-\frac{1}{K}\sum_{k=1}^{K}G_f(r_k),\\
\Delta A_f(v)&=A_f(v)-\frac{1}{K}\sum_{k=1}^{K}A_f(r_k),
\end{aligned}
\label{eq:random-adjusted-context}
\end{equation}
where $A_f$ is the steered-minus-base pair-ranking accuracy change, measured in percentage points. Bootstrap replicates resample pair IDs within contrast and recompute both the adjusted mean and sample standard deviation before aggregation. Positive-pair and same-sign rates are sign summaries only. The main pattern is stable across the five random-control draws: mean standardized effects (seed SD; range) are 4.94 (.10; 4.84--5.06), $-.00$ (.13; $-.17$--.18), 5.49 (.18; 5.23--5.70), and 4.71 (.20; 4.46--4.87) for original, reversed, direct, and opaque answers. All 12 strata are positive for original, direct, and opaque answers; 6--8 of 12 are positive under reversal.

\end{document}